\def\confName{CVPR}
\def\confYear{2026}

\def\paperTitle{SHARP: Short-Window Streaming for Accurate and Robust Prediction in Motion Forecasting}

\def\authorBlock{
    Alexander Prutsch \qquad
    Christian Fruhwirth-Reisinger \qquad
    David Schinagl \qquad
    Horst Possegger\\
    Institute of Visual Computing, Graz University of Technology\\
    {\tt\small \{alexander.prutsch, reisinger, david.schinagl, possegger\}@tugraz.at}
}

\newif\ifreview 
\newif\ifarxiv \newcommand{\arxiv}{\arxivtrue}
\newif\ifcamera 
\newif\ifrebuttal 

\arxiv%

\pdfoutput=1
\documentclass[10pt,twocolumn,letterpaper]{article}
\ifreview \usepackage[review]{cvpr} \fi
\ifarxiv \usepackage[pagenumbers]{cvpr} \fi
\ifrebuttal \usepackage[rebuttal]{cvpr} \fi
\ifcamera \usepackage{cvpr} \fi

\usepackage{graphicx}	
\usepackage{amsmath}	
\usepackage{amssymb}	
\usepackage{booktabs}
\usepackage{times}
\usepackage{microtype}
\usepackage{epsfig}
\usepackage{caption}
\usepackage{float}
\usepackage{placeins}
\usepackage{color, colortbl}
\usepackage{stfloats}
\usepackage{enumitem}
\usepackage{tabularx}
\usepackage{xstring}
\usepackage{multirow}
\usepackage{xspace}
\usepackage{url}
\usepackage{subcaption}
\usepackage{xcolor}
\usepackage[hang,flushmargin]{footmisc}
\usepackage{pifont}
\usepackage{tikz}
\usetikzlibrary[positioning, fit]

\ifcamera \usepackage[accsupp]{axessibility} \fi

\newlength\secondblock %
\newlength\visheight   %

\newcommand{\nbf}[1]{{\noindent \textbf{#1.}}}

\newcommand{\senc}{{\text{enc}}}
\newcommand{\encmodequery}{\ensuremath{Q^t_\senc}}
\newcommand{\encagent}{\ensuremath{A^t_\senc}}
\newcommand{\enclane}{\ensuremath{L^t_\senc}}

\newcommand{\bval}[1]{\textbf{#1}}
\newcommand{\bvc}[1]{\cc \textbf{#1}}
\newcommand{\sval}[1]{\underline{#1}}

\ifarxiv  \fi

\definecolor{rcol}{rgb}{0.9,0.9,0.9}
\newcommand{\cc}[0]{\cellcolor{rcol}}

\newcommand{\R}[1]{{%
    \textbf{%
        \ifstrequal{#1}{BR}{\textcolor{teal}{karx}}{%
        \ifstrequal{#1}{BA}{\textcolor{blue}{U4cY}}{%
        \ifstrequal{#1}{WA}{\textcolor{olive}{ywaZ}}{%
        \ifstrequal{#1}{4}{\textcolor{teal}{R#1}}{%
                           \textcolor{cyan}{R#1}%
        }}}}%
    }%
}}

\newcommand{\mn}{SHARP}
\newcommand{\cmark}{\ding{51}}%
\newcommand{\xmark}{\color{gray}{\ding{55}}}
\newcommand{\pt}{$t_p$}
\newcommand{\cl}{$T_\text{cl}$}
\newcommand{\fh}{$T_f$}

\newcommand{\addresrow}[9]{

    \edef\temp{#9}

    \ifdefempty{\temp}{
        \node (#1_0) {\includegraphics[trim={#3cm, #4cm, #5cm, #6cm}, clip, width=7cm]{figures/raw_figures/results/sharp_#2_2.pdf}};
    }{
        \node (#1_0) [below=of #9_0, yshift=-0.9cm] {\includegraphics[trim={#3cm, #4cm, #5cm, #6cm}, clip, width=7cm]{figures/raw_figures/results/sharp_#2_2.pdf}};
    }

    \node (#1_1) [right=of #1_0, xshift=#7*7cm] {\includegraphics[trim={#3cm, #4cm, #5cm, #6cm}, clip, width=7cm]{figures/raw_figures/results/sharp_#2_3.pdf}};
    \node (#1_2) [right=of #1_1, xshift=#7*7cm] {\includegraphics[trim={#3cm, #4cm, #5cm, #6cm}, clip, width=7cm]{figures/raw_figures/results/sharp_#2_4.pdf}};
    
    \node (#1_3) [right=of #1_2, xshift=#7*7+0.6cm] {\includegraphics[trim={#3cm, #4cm, #5cm, #6cm}, clip, width=7cm]{figures/raw_figures/results/demo_#2_4.pdf}};

    \ifdefempty{\temp}{
        \draw[thick, gray] ([xshift=0.3cm]#1_2.south east) -- ([xshift=0.3cm, yshift=0.5cm]#1_2.north east);
    }{
        \draw[thick, gray] ([xshift=0.3cm]#1_2.south east) -- ([xshift=0.3cm]#9_2.south east);
    }
    
    \newcount\mycount
    \mycount=0

    \newread\myfile
    \openin\myfile=figures/raw_figures/results/#2.txt
    \loop
        \unless\ifeof\myfile
        \read\myfile to \linecontent
        \ifnum\mycount<4
            \node [above=of #1_\the\mycount, yshift=#8*7cm] {$\linecontent$};
        \fi
        \advance\mycount by 1
    \repeat
    \closein\myfile
}

\newcommand{\addfailrow}[9]{
    \edef\temp{#9}
    
    \ifdefempty{\temp}{
        \node (#1_0) {\includegraphics[trim={#3cm, #4cm, #5cm, #6cm}, clip, width=7cm]{figures/raw_figures/fail/sharp_#2_2.pdf}};
    }{
        \node (#1_0) [below=of #9_0, yshift=-0.9cm] {\includegraphics[trim={#3cm, #4cm, #5cm, #6cm}, clip, width=7cm]{figures/raw_figures/fail/sharp_#2_2.pdf}};
    }

    \node (#1_1) [right=of #1_0, xshift=#7*7cm] {\includegraphics[trim={#3cm, #4cm, #5cm, #6cm}, clip, width=7cm]{figures/raw_figures/fail/sharp_#2_3.pdf}};
    \node (#1_2) [right=of #1_1, xshift=#7*7cm] {\includegraphics[trim={#3cm, #4cm, #5cm, #6cm}, clip, width=7cm]{figures/raw_figures/fail/sharp_#2_4.pdf}};
    \node (#1_3) [right=of #1_2, xshift=#7*7+0.6cm] {\includegraphics[trim={#3cm, #4cm, #5cm, #6cm}, clip, width=7cm]{figures/raw_figures/fail/demo_#2_4.pdf}};

    \node (#1_text) at ([xshift=-0.5cm]#1_0.west) [anchor=south, rotate=90] {#8};
    
    \ifdefempty{\temp}{
        \draw[thick, gray] ([xshift=0.3cm]#1_2.south east) -- ([xshift=0.3cm, yshift=0.5cm]#1_2.north east);
    }{
        \draw[thick, gray] ([xshift=0.3cm]#1_2.south east) -- ([xshift=0.3cm]#9_2.south east);
    }
    
    \newcount\mycount
    \mycount=0

    \openin\myfile=figures/raw_figures/fail/#2.txt
    \loop
        \unless\ifeof\myfile
        \read\myfile to \linecontent
        \ifnum\mycount<4
            \node [above=of #1_\the\mycount, yshift=0.01*7cm] {$\linecontent$};
        \fi
        \advance\mycount by 1
    \repeat
    \closein\myfile
}

\usepackage{xr-hyper}

\makeatletter
\newcommand*{\addFileDependency}[1]{
  \typeout{(#1)}
  \@addtofilelist{#1}
  \IfFileExists{#1}{}{\typeout{No file #1.}}
}

\makeatother
\newcommand*{\myexternaldocument}[1]{
    \externaldocument{#1}
    \addFileDependency{#1.tex}
    \addFileDependency{#1.aux}
}

\definecolor{cvprblue}{rgb}{0.21,0.49,0.74}
\usepackage[pagebackref,breaklinks,colorlinks,allcolors=cvprblue]{hyperref}
\usepackage[capitalize]{cleveref}
\crefname{section}{Sec.}{Secs.}
\crefname{table}{Table}{Tables}
\crefname{figure}{Fig.}{Figs.}

\ifarxiv \crefname{appendix}{App.}{Apps.}
\else \crefname{appendix}{Suppl.}{Suppls.} \fi

\unless\ifarxiv \myexternaldocument{_supplementary} \fi

\begin{document}
\title{\paperTitle}
\author{\authorBlock}
\maketitle

\begin{abstract}
In dynamic traffic environments, motion forecasting models must be able to accurately estimate future trajectories continuously.
Streaming-based methods are a promising solution, but despite recent advances, their performance often degrades when exposed to heterogeneous observation lengths.
To address this, we propose a novel streaming-based motion forecasting framework that explicitly focuses on evolving scenes.
Our method incrementally processes incoming observation windows and leverages an instance-aware context streaming to maintain and update latent agent representations across inference steps.
A dual training objective further enables consistent forecasting accuracy across diverse observation horizons.
Extensive experiments on Argoverse 2, nuScenes, and Argoverse 1 demonstrate the robustness of our approach under evolving scene conditions and also on the single-agent benchmarks.
Our model achieves state-of-the-art performance in streaming inference on the Argoverse 2 multi-agent benchmark, while maintaining minimal latency, highlighting its suitability for real-world deployment.
\end{abstract}

\section{Introduction}
\label{sec:intro}
\begin{figure}[tp]
    \newlength{\imgh} 
    \settoheight{\imgh}{\includegraphics{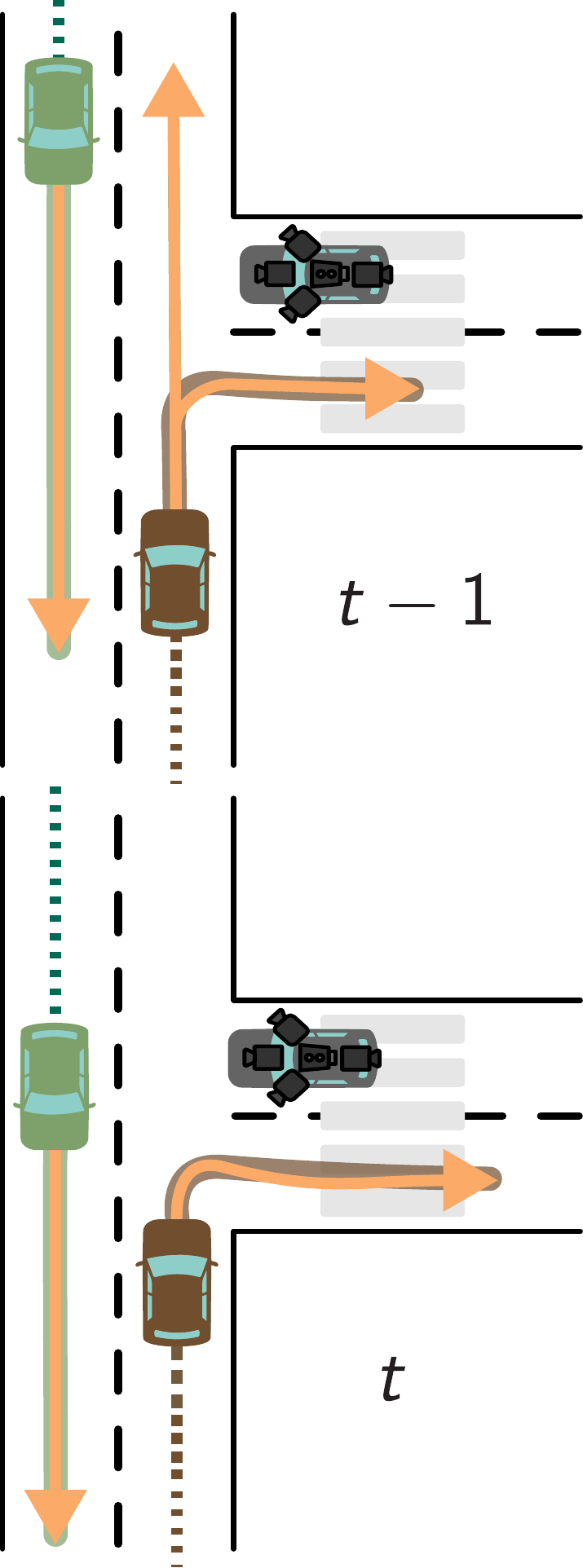}}

    \begin{tabular}{c@{\hspace{0.02\linewidth}}c} 
        \includegraphics[page=1, width=0.48\linewidth]{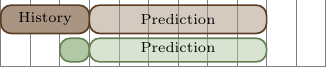} &
        \includegraphics[page=2, width=0.48\linewidth]{figures/raw_figures/streaming_windows/teaser.pdf} \\

        \includegraphics[trim={0cm 0.5\imgh{}cm 0cm 0cm}, clip, width=0.4\linewidth]{figures/raw_figures/_teaser.pdf} &
        \includegraphics[trim=0cm 0cm 0cm {0.5\imgh{}cm}, clip, width=0.4\linewidth]{figures/raw_figures/_teaser.pdf} \\
    \end{tabular}
    
    \vspace{-0.3cm}
    \caption{
    In real-world driving scenes, the available context history for different agents is heterogeneous.
    Therefore, forecasting models should be able to make accurate predictions for both long-term (brown car) and short-term (green car) contexts, where agents have been observed for a long time or have recently entered the field of view.
    While existing models often struggle with such varying context lengths, our \mn~is explicitly designed to provide accurate forecasts in such dynamically evolving scenes.
    Dotted lines indicate the available observations, while transparent paths show the corresponding future trajectory.
    }
    \label{fig:teaser}
    \vspace{-0.5cm}
\end{figure}

Trajectory prediction is a core component of the autonomous vehicle (AV) control stack, providing hypotheses on the future motions of surrounding agents based on perception outputs, typically detections and tracks.
Accurate and robust predictions are critical for safe and efficient motion planning which enables the AV to anticipate and respond to dynamic behaviors in complex traffic scenarios.
In real-world driving, traffic scenes are constantly changing, as illustrated in \Cref{fig:teaser}: Agents entering the field-of-view of the AV have been observed only briefly, whereas for other agents a more comprehensive motion history is available.
Motion forecasting models must therefore be able to \textit{leverage heterogeneous historical observations} effectively while operating under real-time constraints in continuously evolving scenes.

Trajectory prediction has been extensively studied~\cite{shi2022motion, zhou2023query, nayakanti2023wayformer, cheng2023forecast, zhang2024demo}, with state-of-the-art methods achieving high accuracy on large-scale datasets~\cite{chang2019argoverse, caesar2020nuscenes, wilson2021argoverse, ettinger2021large}.
However, these benchmarks consider only fixed-size historical and future windows, while in practice, the historical context can range from a few frames up to several seconds.
Methods that rely on extensive contexts typically achieve the most accurate results, but must delay predictions until a sufficient amount of observations are accumulated for newly detected agents.
Only a few works, \eg~\cite{xu2024adapting, li2024lakd}, evaluate varying observation lengths, usually by partially masking agent histories.
To handle the constantly evolving context of real-world traffic scenes, however, we require models that can efficiently propagate motion features as long as the agents are visible.
This challenge has not yet been sufficiently addressed.

To enable reliable motion forecasting under heterogeneous observation lengths, we propose a novel trajectory prediction framework and training scheme.
Our model, \mn~(\textit{\underline{sh}ort-window streaming for \underline{a}ccurate and \underline{r}obust \underline{p}redictions)}, operates in a streaming mode and processes short observation windows while effectively propagating context information across multiple time steps.
We introduce a new instance-aware context streaming module and jointly optimize for both long-context and single-chunk prediction.

Our \mn~employs an efficient transformer-based architecture that demonstrates strong robustness across varying context lengths on the Argoverse~2 (AV2)~\cite{wilson2021argoverse} dataset, which features long and complex driving scenarios.
We conduct a comprehensive evaluation using different context lengths to assess performance across diverse temporal settings.
Additional experiments on Argoverse~1~(AV1)~\cite{chang2019argoverse} and nuScenes~\cite{caesar2020nuscenes} further demonstrate the generality and effectiveness of our method.
{Compared to prior streaming methods, \eg~\cite{song2024realmotion, zhang2024demo}, which relay information but remain constrained to fixed-size training inputs, our design generalizes more effectively to varying lengths.
In contrast to models trained on variable histories, \eg~\cite{xu2024adapting}, \mn~incorporates the gradual evolution of traffic scenes which leads to favorable performance under varying observation lengths.
}
Our method matches state-of-the-art performance in the standard single-agent evaluation setting and establishes a new state-of-the-art for streaming multi-agent prediction.
\noindent In summary, our main contributions are:
\begin{itemize}
    \item We propose \mn, a novel trajectory forecasting approach, effectively predicting accurate trajectories on various observation lengths in continuously evolving scenes, tackling key challenges in real-world autonomous driving.
    \item Extensive evaluations on AV2, nuScenes, and AV1 show that our method excels across various datasets providing a lightweight yet effective motion forecasting model.
    \item We demonstrate robustness across a wide range of variable observation lengths, a highly relevant evaluation that should be thoroughly addressed by our community.
\end{itemize}

\section{Related Work}
\label{sec:related}
State-of-the-art trajectory prediction models typically encode historical trajectories of agents and map information as a first processing step.
While historical agent tracks are commonly encoded using temporal self-attention~\cite{lan2023sept, cheng2023forecast, prutsch24efficient, song2024realmotion} or state-space models~\cite{zhang2024demo, huang2025trajectory}, map information is usually represented in vectorized form~\cite{gao2020vectornet} and encoded via PointNet-like architectures~\cite{qi2017pointnet}.
To model relationships between all scene elements, individual encodings are combined either with attention blocks~\cite{liu2021multimodal, ngiam2021scene, shi2022motion, shi2024mtr++, nayakanti2023wayformer, zhang2023hptr, gan2024mgtr} or graph neural networks~\cite{gao2020vectornet, jia2023hdgt, zhou2023query, cui2023gorela, tang2024hpnet}.
DETR-based architectures~\cite{carion2020end} have shown strong performance~\cite{shi2022motion, zhou2023query, prutsch24efficient} for decoding multi-modal future trajectories.
Recent advances in trajectory prediction focus on improving various aspects, like scene modeling~\cite{zhou2023query}, refinement decoders~\cite{zhou2024smartrefine}, or improved model architectures~\cite{zhang2024demo, huang2025trajectory}.
While existing methods achieve high accuracy on benchmark datasets~\cite{chang2019argoverse, caesar2020nuscenes, wilson2021argoverse, ettinger2021large}, they largely rely on fixed-size inputs and overlook real-world challenges such as varying observation lengths and continuous operation in evolving scenes.
Our approach explicitly addresses these aspects by emphasizing flexible input sizes, efficiency, and temporal consistency.

\nbf{Trajectory Prediction with Varying Observation Lengths}
\emph{FlexiLength Network~(FLN)}~\cite{xu2024adapting} presents a method for handling observation lengths shifts using two components: \emph{\mbox{FlexiLength} Calibration~(FLC)}, enabling an observation length invariant feature representation, and \emph{FlexiLength Adaptation~(FLA)}, which then adapts to specific observation lengths.
POP~\cite{wang2024improving} uses self-supervised learning for mask history reconstruction as a pretext task to handle dynamic observation lengths. 
Feature distillation is then used to transfer knowledge from a fully observed teacher model to a student model with limited observations.
LaKD~\cite{li2024lakd} introduces a dynamic knowledge distillation method where a single encoder, acting as both teacher and student, adaptively transfers knowledge across trajectories of different lengths.
It also employs a dynamic soft-masking mechanism to mitigate knowledge conflicts during this self-distillation process.

Processing sequences of varying input lengths offers greater flexibility compared to models trained on fixed-size inputs.
In real-world driving, however, models must handle streams of observations that gradually extend over time. 
While existing approaches perform well across different sequence lengths, they lack information-streaming mechanisms, resulting in consecutive predictions that are independent.
We address this limitation by processing sequences in chunks, mirroring the progressive nature of real-world inputs, while introducing a streaming mechanism that promotes temporal continuity and captures long-term dependencies.

\nbf{Streaming Trajectory Prediction}
RealMotion~\cite{song2024realmotion} introduces two information streaming mechanisms for motion forecasting in continuous settings.
Unlike traditional snapshot-based methods~\cite{shi2022motion, zhou2024smartrefine, huang2025trajectory}, which predict each frame independently, these mechanisms leverage information across multiple observation windows, enabling more consistent predictions.
Specifically, a cross-attention-based scene context streamer integrates context of the previous step into the current scene.
A motion-aware layer normalization (MLN)~\cite{wang2023exploring} compensates the coordinate frame displacement. 
In addition, a trajectory relay mechanism incorporates past predictions into the current inference.
DeMo~\cite{zhang2024demo} adopts the aforementioned streaming mechanisms, achieving state-of-the-art results on the Argoverse~2 dataset.

While RealMotion provides a promising framework for continuous motion forecasting, it has several limitations.
First, its end-to-end training scheme is not well suited to variable sequence lengths, as the models are optimized on a fixed number of streaming passes, limiting flexibility across different streaming steps.
Second, each inference pass uses a large 3\,s observation window, which introduces delays for newly detected agents.
Finally, although the context relay mechanism is promising, it does not explicitly model instance correspondences during context streaming and therefore captures long-term relationships only implicitly.
In contrast, our approach maintains consistent agent encodings across context updates, employs short input windows, and is specifically trained to handle varying observation lengths.

\begin{figure*}[tp]
    \centering
    \includegraphics[trim={0cm 0.7cm 0cm 0cm}, clip, width=\linewidth]{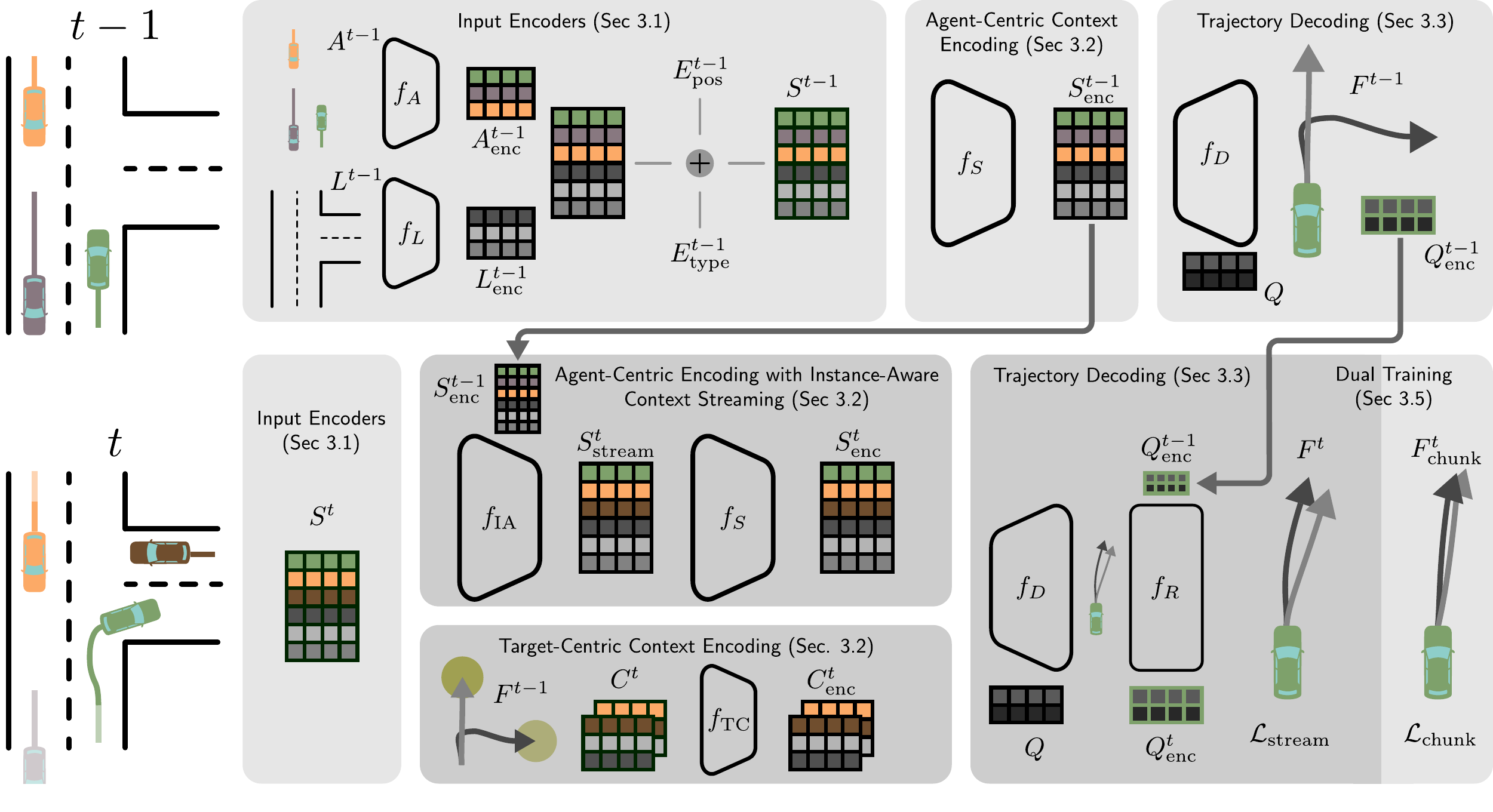} 
    \vspace{-0.4cm}
    \caption{
    To jointly incorporate newly detected agents and long-term agent histories when forecasting motions in evolving scenes, we leverage a streaming-based motion forecasting model.
    The example at time step $t-1$ illustrates the standard model pass without streaming context, consisting of separate agent and lane encoders ($f_A$ and $f_L$), a scene encoder ($f_S$), and a trajectory decoder ($f_D$) with a streaming refinement module ($f_R$).
    At the next time step $t$, we integrate the previous scene context $S_\senc^{t-1}$ via an instance-aware context streamer ($f_{\text{IA}}$) and generate auxiliary target-centric features $C^t$ by aggregating scene elements closely around the endpoints of the previous predictions $F^{t-1}$.
    To improve robustness, we also perform, only during training, a parallel model pass without streaming modules, producing $F^t_{\text{chunk}}$ which is used to compute the $\mathcal{L}_\text{chunk}$ objective.
    }
    \label{fig:arch}
    \vspace{-0.3cm}
\end{figure*}

\section{SHARP Motion Forecasting}
\label{sec:method}
We propose a novel trajectory prediction architecture and training scheme to address three key challenges:
(1) achieving accurate predictions from short-horizon observation histories, (2) enabling streaming information propagation for operation in continuously evolving scenes, and (3) maintaining a lightweight architecture suitable for real-time inference.
Our approach builds on an efficient transformer-based backbone and avoids extra bells and whistles, such as iterative refinement during decoding~\cite{shi2022motion, zhou2023query, zhou2024smartrefine}, which typically provide only marginal performance improvements while adding substantial latency.
\Cref{fig:arch} provides an overview of our architecture.
We first process historical agent data and map information to construct a scene representation (\cref{sec:ale}).
As new agent observations arrive, our model incrementally updates this representation by fusing the incoming states with the previous scene context.
To propagate information reliably over time, we introduce an instance-aware context streamer~(\cref{sec:ssce}) that explicitly models agent correspondences.
We further enhance the context representation using the endpoints from previous predictions to extract target-centric auxiliary features.
A DETR-like decoder (\cref{sec:td}) is utilized to predict the trajectories, which can be easily extended for consistent predictions of all agents in the scene (\cref{sec:etswjp}). 
During training (\cref{sec:dpts}), we jointly optimize the model to predict trajectories from both the streamed scene context and the single observation context, promoting more robust representations across varying observation lengths.

\nbf{Problem Definition}
In autonomous driving, motion forecasting aims to provide estimates $F$ for an agent’s future trajectories $F$ alongside their associated probability scores $P$, conditioned on the historical states $A$ of all observed agents and the surrounding map information $L$.
Following common practice~\cite{cheng2023forecast, prutsch24efficient}, we sample all agents and map elements within a fixed-radius region centered on the focal agent.
The agent states are represented as a tensor $A\in \mathbb{R}^{N_a \times T_h \times D_a}$ , where $N_a$ denotes the number of agents, $T_h$ the number of state observations and $D_a$ the feature dimension, \ie historical positions and headings.
Similarly, the map is represented as a tensor $L\in \mathbb{R}^{N_l \times P_l \times D_l}$, where $N_l$ is the number of lane segments, $P_l$ the number of sampled points per segment, and $D_l$ the feature dimension (typically the $xy$-coordinates).
Segments are defined by uniformly sampling $P_l$ points along the lane centerline.

The trajectory prediction output consists of multi-modal future trajectories $F \in \mathbb{R}^{K \times T_f \times D_f}$ and their corresponding probability scores $P \in \mathbb{R}^{K}$, where $K$ is the number of motion modes (hypotheses), $T_f$ the prediction horizon (number of future time steps), and $D_f$ is the feature dimension (typically, $D_f=2$, \ie $xy$-coordinates).
At a given time step~$t$, the non-streaming (\emph{snapshot-based}) trajectory prediction process can be expressed as $(F^t, P^t) = f(A^t, L^t)$, where $f$ denotes the trajectory prediction model including an encoder $f_E$ and a decoder $f_D$.
In our approach, the encoder $f_E$ creates a standard agent-centric scene context $S_\senc^t$ and auxiliary target-centric features $C_\senc^t$ (details in \Cref{{sec:tcce}}).
The decoder $f_D$ then predicts the output using both contexts: $(F^t, P^t)=f_D(S_\senc^t, C_\senc^t)$.
To address the evolving nature of real-world scenarios, we extend our model to a streaming formulation.
The total available context length is denoted by \cl, which the model processes through multiple streaming passes using an observation window of size $T_h$.
We leverage the temporal continuity between consecutive passes by incorporating the previous scene context $S_\senc^{t-1}$ and prior predictions $F^{t-1}$ into the current prediction step.
This leads to the updated process $(F^t, P^t) = f(A^t, L^t, S_\senc^{t-1}, F^{t-1})$, where $S_\senc^{t-1}$ is integrated during encoding and $F^{t-1}$ is fused at the final stage during decoding.

\subsection{Agent and Lane Encoding}
\label{sec:ale}
We first encode all agent motions and lane geometries using element-wise local coordinates~\cite{cheng2023forecast}.
This improves learning efficiency, as motion patterns and lane shapes are modeled independently of their global positions.
Agent states are normalized \wrt each agent’s most recent pose, while lane segments are expressed relative to their polyline center pose.
For agent encoding, we employ a compact self-attention-based encoder $f_A$ to process the historical agent states $A^t$~\cite{lan2023sept, prutsch24efficient}.
The agent states are first projected into a $D$-dimensional feature space via a linear layer.
Subsequently, multi-head self-attention blocks~\cite{vaswani2017attention} capture agent motion dynamics, followed by a max-pooling operation across the temporal dimension.
The resulting agent embeddings are given by $\encagent=f_A(A^t) \in \mathbb{R}^{N_a \times D}$.
To encode the lane information $L^t$, we adopt a PointNet-based~\cite{qi2017pointnet} lane encoder~$f_L$, following ~\cite{cheng2023forecast, song2024realmotion, zhang2024demo}.
This results in a lane context representation $\enclane = f_L(L^t) \in \mathbb{R}^{N_l \times D}$.

\subsection{Streaming Scene Context Encoding}
\label{sec:ssce}
Following the individual encoding of scene elements in local coordinates, we concatenate the agent $\encagent$ and lane $\enclane$  tokens to form a scene representation $\mathbb{R}^{(N_a+N_l) \times D}$.
We then establish a global scene representation relative to the focal agent’s pose by augmenting each token with global positional embeddings derived from the current pose of each agent and the center pose of each lane segment~\cite{cheng2023forecast}.
Each pose is defined by its position ($x, y$) and heading/rotation ($\gamma$), which we encode as $(x, y, \sin \gamma, \cos \gamma)$.
These pose features are processed by a two-layer multilayer perceptron (MLP) to produce positional embeddings $E_{\text{pos}}^t \in \mathbb{R}^{(N_a+N_l) \times D}$.
Additionally, we include a type embedding $E_{\text{type}}^t \in \mathbb{R}^{(N_a+N_l) \times D}$ for each agent and lane token to capture categorical information such as the agent class or lane type.
This yields a scene representation $S^t \in \mathbb{R}^{(N_a+N_l) \times D}$.

\nbf{Instance-Aware Context Streaming}
Our model is designed to operate in a streaming setting, where scene information evolves continuously over time.
At each time step $t$, the motion of all agents and the nearby map information for the current observation window is encoded in the scene representation $S^t$.
To enable temporal information propagation, we incorporate the past scene context $S_\senc^{t-1}$ into the current representation $S^t$.
Particularly, we introduce a cross–attention–based context streaming module $f_\text{IA}$ that explicitly leverages the temporal correspondences of an agent's encoding throughout its trajectory. 
Given the instance information inherently provided by the input trajectory, we implement instance-aware context streaming by leveraging an attention mask.
This mask selectively modulates cross-attention weights, steering focus toward matching instances via a learned bias parameter.
In contrast, prior context-streaming approaches~\cite{song2024realmotion, zhang2024demo} rely solely on positional correspondences, modeling instance relations only implicitly within the streaming module despite their explicit use in constructing the agent encoder inputs $A^t$.
Our attention mask-based design enables the model to incorporate these explicit correspondences while preserving spatial relational reasoning.
Moreover, unlike conventional methods that assume perfect agent associations over extended context windows $T_{\hat{\text{cl}}}$, our short-window design intrinsically enables recovery from tracking discontinuities.
The context streaming mechanism generates an enhanced scene context $S_\text{stream}^t$, which effectively integrates and leverages information from preceding observation windows.

\nbf{Agent-Centric Context Encoding}
After integrating the previous context $S_\senc^{t-1}$, we use a self-attention-based scene encoder $f_S$~\cite{cheng2023forecast, lan2023sept} to capture relationships between scene elements, including agent–agent interactions, map topology, and agent–map interactions.
This yields a learned scene context $S_\senc^t = f_S(S_\text{stream}^t) \in \mathbb{R}^{(N_a+N_l) \times D}$, which encodes not only the individual observations but also the relations learned among all elements.

\nbf{Target-Centric Context Encoding}
\label{sec:tcce}
In addition to the standard agent-centric context, we incorporate a set of target-centric features~\cite{zhou2024smartrefine, wang2023ganet, prutsch2026streaming} for enhanced performance in our streaming setting.
Specifically, we leverage the $K$ trajectory endpoints from the previous inference step, which represent map regions of potentially high relevance, as anchors to extract additional contextual features $C^t$~\cite{prutsch2026streaming}.
Each $C^t$ is obtained by aggregating all tokens within a compact region of interest centered at the corresponding endpoint.
The target-centric context encoder $f_\text{TC}$ adopts the same architecture as the agent-centric encoder $f_S$, while the reduced token set ensures low latency overhead.
We define the endpoints as the origin of their respective local coordinate systems to model the target-centric features effectively.
To maintain global geometric relationships, we further incorporate a positional embedding that explicitly encodes the spatial relationship between the focal agent and each endpoint.
The result is given as $C_\senc^t \in \mathbb{R}^{K \times (N_a+N_l)' \times D}$ where $(N_a+N_l)'$ is the maximum number of tokens in the reduced sets.

\subsection{Trajectory Decoding}
\label{sec:td}
We employ a DETR-like~\cite{carion2020end} decoder to generate $K$ multi-modal future trajectories $F^t$ along with their corresponding probability scores $P^t$.
Our decoder $f_D$ follows the standard procedure, applying cross-attention between learnable mode queries $Q \in \mathbb{R}^{K \times D}$ and the scene features $S_\senc^t$ and $C_\senc^t$.
To capture both global and target-specific context, we use separate cross-attention blocks for agent-centric and target-centric features.
Finally, two shallow MLP heads decode the updated queries $\encmodequery=f_\text{D}\left( S_\senc^t, C_\senc^t, Q \right)$ into the future trajectories and their associated probabilities.

\nbf{Trajectory Relay}
To enhance prediction consistency across streaming inference steps, we adopt a trajectory relay mechanism~\cite{song2024realmotion}.
Previous forecasts are transformed into the current coordinate frame and used to refine the current predictions via a cross-attention module $f_R$.
This design enables the model to refine its outputs based on past information without being constrained by previous predictions, for example, when new observations render earlier forecasts unlikely.

\subsection{Extension to Scene-wide Joint Predictions}
\label{sec:etswjp}
To obtain scene-consistent multi-agent predictions (joint predictions), we combine single-agent predictions (marginal predictions) by leveraging the encoded mode queries \encmodequery.
Initially, we execute marginal predictions for each relevant agent in the traffic scene to obtain the individual mode queries \encmodequery.
Subsequently, we augment these queries with positional embeddings of the agents relative to a scene-global coordinate system, \ie relative to a reference agent or the ego vehicle.
We then apply an interaction block, consisting of cross-attention across all agents within the scenario and cross-attention across all modes per agent. 
This explicitly models the spatial relationships and interactions between agents---considering multiple possible trajectories for each---and yields $K$ joint world predictions for the motion of all agents.
The final trajectories are decoded using the same head structure as employed for the single-agent case.

\subsection{Dual Training}
\label{sec:dpts}
We propose a dual training loss to enhance robustness across varying observation windows, rather than optimizing solely for a fixed context length.
In contrast to prior streaming approaches~\cite{song2024realmotion}, we reduce each streaming step to a short observation window (\eg 1\,s).
This is aligned with the concept of gradually acquiring new observations over time.
Additionally, this design enables more frequent gradient updates within each training scenario and supports longer prediction horizons, as only a short history is required for the initial training pass.
During training, we employ non-overlapping windowing to balance input diversity and efficiency.
For every new observation chunk, the model predicts future trajectories both with and without streaming context.
We compute separate losses for each case: $\mathcal{L}_{\text{stream}}$ for predictions $F^t$ conditioned on the accumulated streaming history, and $\mathcal{L}_{\text{chunk}}$ for predictions $F^t_{\text{chunk}}$ based solely on the current observation chunk.
Following prior work~\cite{cheng2023forecast, prutsch24efficient, song2024realmotion, zhang2024demo}, each term combines a cross-entropy classification loss to assign the highest probability to the best-fitting trajectory and a Smooth L1 regression loss~\cite{huber1964robust} to fit the predicted trajectories to the ground truth, under a winner-takes-all strategy where only the trajectory with the smallest displacement error contributes to the gradient.
The final objective is given by $\mathcal{L}_{\text{dual}}=\mathcal{L}_{\text{stream}} + \mathcal{L}_{\text{chunk}}$.

\section{Experiments}
We evaluate our approach using three large-scale autonomous driving datasets: Argoverse~1~(AV1)~\cite{chang2019argoverse}, Argoverse~2~(AV2)~\cite{wilson2021argoverse} (single- and multi-agent settings), and nuScenes~\cite{caesar2020nuscenes}.
Additional model ablations, robustness evaluations, as well as full implementation details and training protocol are provided in the supplementary material.

\subsection{Experimental Settings}

\begin{figure}[tp]
    \centering
        \includegraphics[trim={0cm, 0cm, 0cm, 0cm}, clip, width=0.95\linewidth]{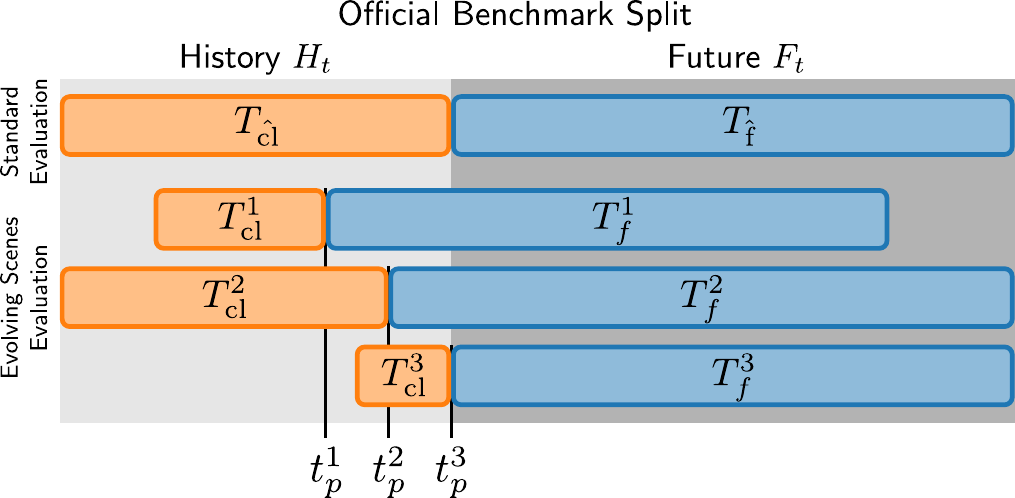} 
    \vspace{-0.2cm}
    \caption{
    Comparison of different evaluation setups on a motion forecasting dataset with $H_t$ as historical context and $F_t$ as future for the prediction task.
    The standard benchmark evaluation only considers a single history/future split per scenario (first row).
    In our experiments, we test the models at different time steps $t_p^i$ into the scenario and using varying context lengths $T^i_{\text{cl}}$ as model input and also evaluating for different future horizons $T^i_f$.
    }
    \label{fig:timeline}
    \vspace{-0.45cm}
\end{figure}

\nbf{Datasets}
Both Argoverse datasets~\cite{chang2019argoverse, wilson2021argoverse} were collected in the United States and recorded at 10\,Hz.
AV1~\cite{chang2019argoverse} includes 324k scenarios, each 5\,s long, with $H_t$=2\,s of historical context and a $F_t$=3\,s prediction horizon.
AV2~\cite{wilson2021argoverse} comprises 250k scenarios with longer sequences of 11\,s (5\,s history, 6\,s future).
The nuScenes prediction dataset~\cite{caesar2020nuscenes} is derived from 850 scenes of the nuScenes dataset and consists of 50k scenarios sampled at only 2\,Hz, each featuring 2\,s of history and 6\,s of future motion.
AV1 and nuScenes adopt a \textit{single-agent evaluation} protocol, focusing on the prediction of \textit{a designated focal agent per scenario}.
In contrast, AV2 additionally supports a \textit{multi-agent evaluation} setting, where predictions for \textit{multiple agents within a scene are jointly assessed}, which closely reflects the requirements of real-world applications.
Given its longer temporal span, higher frame rate, and multi-agent setting, AV2 serves as our primary benchmark for streaming-based evaluation.
We further employ AV1 and nuScenes to ensure comparability with prior work and to conduct additional ablation studies.

\nbf{Metrics}
We evaluate our methods following the standard benchmark protocols using the miss rate (MR$_k$), minimum average displacement error (minADE$_k$), minimum final displacement error (minFDE$_k$), and brier-minFDE$_6$ as metrics.
For each metric, the $k$ highest scoring trajectory hypotheses are considered and the minimum errors are reported based on the best-fitting trajectory.
The brier-minFDE$_6$ also includes a penalty for assessing the probability score of the best fitting trajectory.
For the AV2 multi-agent setting, we use the standard metrics which evaluate different \emph{worlds} considering the predictions for each agent in a scene, namely actorMR$_k$, avgMinADE$_k$, avgMinFDE$_k$ and avgBrierMinFDE$_k$.

\nbf{Evolving Scenes Evaluation Setup}
In standard trajectory prediction benchmarks, models are evaluated using fixed input and output windows: a historical context of length $H_t$ and ground-truth future $F_t$.
To obtain a more comprehensive and realistic assessment, we introduce an evaluation protocol that emulates continuously evolving scenes with heterogeneous observation lengths (see \Cref{fig:timeline}).
Within this framework, streaming-based methods are evaluated on varying context lengths \cl~at different inference time steps \pt~which also affect the available prediction horizons \fh.
The conventional evaluation protocol is recovered when \cl = $H_t$, \pt = $H_t$, and \fh = $F_t$.

\newcommand{\pl}[0]{\multicolumn{3}{c}{\cellcolor{white}\phantom{a}}}
\newcommand{\win}[1]{\multirow{3}{*}{\includegraphics[page=#1]{figures/raw_figures/streaming_windows/evolving_scenes_table.pdf}}}

\begin{table*}[t]
\footnotesize
\centering
\setlength{\tabcolsep}{6pt}
\begin{tabular}{cc|cccccccc}
\toprule
History / Prediction Time Step / Future Horizon & Method  & minADE$_1$ & minFDE$_1$ & MR$_6$ & minADE$_6$ & minFDE$_6$ & \textbf{brier-minFDE}$_\mathbf{6}$ \\ 
\midrule
\win{2} & RealMotion~\cite{song2024realmotion} & 1.70 & 4.44 & 0.17 & 0.68 & 1.35 & 2.00  \\
& DeMo~\cite{zhang2024demo} & 1.58 & 4.21 & 0.16 & \bval{0.62} & 1.26 & 1.95 \\ 
& \cc \mn~(Ours)            & \bvc{1.54} & \bvc{3.84} & \bvc{0.13} & \cc 0.63 & \bvc{1.18} & \bvc{1.80} \\ 
&&&&&& \\ [-0.8em]
\win{3} & RealMotion~\cite{song2024realmotion} & 2.04 & 5.33 & 0.23 & 0.79 & 1.65 & 2.32 \\
& DeMo~\cite{zhang2024demo} & 1.92 & 5.16 & 0.25 & 0.79 & 1.78 & 2.48 \\
& \cc \mn~(Ours)               & \bvc{1.84} & \bvc{4.66} & \bvc{0.19} & \bvc{0.73} & \bvc{1.44} & \bvc{2.09} \\
&&&&&& \\ [-0.8em]
\win{4} & RealMotion~\cite{song2024realmotion} & 1.81 & 4.81 & 0.22 & 0.72 & 1.56 & 2.25 \\
& DeMo~\cite{zhang2024demo}            & 1.70 & 4.66 & 0.25 & 0.74 & 1.75 & 2.45 \\
& \cc \mn~(Ours)                           & \bvc{1.58} & \bvc{4.07} & \bvc{0.18} & \bvc{0.66} & \bvc{1.36} & \bvc{2.02} \\ 
\midrule
\win{5} & RealMotion~\cite{song2024realmotion} & 1.37 & 3.46 & 0.10 & 0.56 & 1.05 & 1.70 \\
& DeMo~\cite{zhang2024demo}            & 1.23 & 3.15 & 0.10 & 0.52 & 0.98 & 1.66 \\
& \cc \mn~(Ours)                           & \bvc{1.21} & \bvc{2.93} & \bvc{0.09} & \bvc{0.51} & \bvc{0.94} & \bvc{1.56} \\ 
&&&&&& \\ [-0.8em]
\win{6} & RealMotion~\cite{song2024realmotion} & 1.19 & 3.05 & 0.09 & 0.48 & 0.91 & 1.58  \\
& DeMo~\cite{zhang2024demo}            & 1.07 & 2.77 & 0.11 & 0.48 & 0.93 & 1.62 \\ 
& \cc \mn~(Ours)                           & \bvc{0.89} & \bvc{2.10} & \bvc{0.05} & \bvc{0.39} & \bvc{0.70} & \bvc{1.31} \\ 
&&&&&& \\ [-0.8em]
\win{7} & RealMotion~\cite{song2024realmotion} & 0.94 & 2.36 & 0.07 & 0.39 & 0.70 & 1.39  \\
& DeMo~\cite{zhang2024demo}            & 0.84 & 2.15 & 0.10 & 0.41 & 0.78 & 1.48 \\
& \cc \mn~(Ours)                           & \bvc{0.60} & \bvc{1.36} & \bvc{0.02} & \bvc{0.28} & \bvc{0.48} & \bvc{1.08} \\
\midrule
\win{8} & RealMotion~\cite{song2024realmotion} & 1.65 & 4.31 & 0.17 & 0.67 & 1.34 & 1.99 \\
& DeMo~\cite{zhang2024demo}            & \bval{1.54} & 4.11 & 0.15 & \bval{0.61} & 1.27 & 1.95 \\
& \cc \mn~(Ours)                           & \cc 1.59 & \bvc{4.08} & \bvc{0.14} & \cc 0.64 & \bvc{1.23} & \bvc{1.87} \\
&&&&&& \\ [-0.8em]
\win{9} & RealMotion~\cite{song2024realmotion} & 1.77 & 4.51 & 0.18 & 0.71 & 1.39 & 2.04 \\
& DeMo~\cite{zhang2024demo}            & \bval{1.68} & 4.33 & 0.18 & 0.68 & 1.38 & 2.06 \\
& \cc \mn~(Ours)                           & \cc 1.69 & \bvc{4.16}  & \bvc{0.15} & \bvc{0.67} & \bvc{1.25} & \bvc{1.89} \\
&&&&&& \\ [-0.8em]
\win{10} & RealMotion~\cite{song2024realmotion} & 1.70 & 4.37 & 0.17 & 0.67 & 1.34 & 2.00 \\
& DeMo~\cite{zhang2024demo}            &  1.65 & 4.29 & 0.19 & 0.68 & 1.44 & 2.13\\
& \cc \mn~(Ours)                           &  \bvc{1.60} & \bvc{3.98} & \bvc{0.15} & \bvc{0.60} & \bvc{1.21} & \bvc{1.86} \\
\midrule
\win{11} & RealMotion~\cite{song2024realmotion} & 1.65 & 4.10 & 0.16 & 0.67 & 1.30 & 1.94 \\
& DeMo~\cite{zhang2024demo}            & \bval{1.48} & \bval{3.73} & \bval{0.13} & \bval{0.61} & \bval{1.19} & 1.86 \\
& \cc  \mn~(Ours)                           & \cc 1.57 & \cc 3.85 & \cc 0.14 & \cc 0.64 & \cc 1.20 & \bvc{1.82} \\ 
\bottomrule
\end{tabular}
\vspace{-0.1cm}
\caption{
Evaluation of streaming-based methods on the AV2 validation set at \textbf{varying prediction time steps \pt~and context lengths \cl}.
To ensure a fair comparison, we test related work~\cite{song2024realmotion, zhang2024demo} using their standard streaming configuration (3\,s input windows with 1\,s displacement) and only adjust the number of inference passes to vary \cl.
The top block evaluates short observation lengths, while the second block extends the context to achieve streaming processing beyond the standard 3 iterative steps of~\cite{song2024realmotion, zhang2024demo}.
The third block tests streaming with \cl\,=\,4\,s at different prediction times.
The bottom block corresponds to the standard AV2 evaluation.
For predictions at \pt$>$5\,s, the evaluation is truncated as only 11\,s are available.
Further details on the experiment setup are provided in the supplementary material.
}
\vspace{-0.2cm}
\label{tab:res_av2_evolving}
\end{table*}

\begin{table*}[t]
\footnotesize
\centering
\setlength{\tabcolsep}{6pt}
\begin{tabular}{lc|cccccc}
\toprule
Method & Streaming & avgMinADE$_1$ & avgMinFDE$_1$ & actorMR$_6$ & avgMinADE$_6$ & avgMinFDE$_6$ & \textbf{avgBrierMinFDE}$_\mathbf{6}$ \\ 
\midrule
FJMP~\cite{rowe2023fjmp}              & \xmark & 1.52 & 4.00 & 0.23 & 0.81 & 1.89 & 2.59 \\
Forecast-MAE~\cite{cheng2023forecast} & \xmark & 1.30 & 3.33 & 0.19 & 0.69 & 1.55 & 2.24 \\
SRefiner~\cite{xiao2025srefiner}      & \xmark & 1.48 & 3.85 & 0.19 & 0.68 & 1.52 & 2.21 \\
RealMotion~\cite{song2024realmotion}  & \cmark & 1.14 & 2.87 & 0.18 & 0.62 & 1.32 & 2.01 \\
DeMo~\cite{zhang2024demo}             & \cmark & \sval{1.12} & \sval{2.78} & \sval{0.16} & \sval{0.58} & \sval{1.24} & \sval{1.93} \\
\rowcolor{rcol}\mn~(Ours)             & \cmark & \bval{1.03} & \bval{2.53} & \bval{0.15} & \bval{0.56} & \bval{1.15} & \bval{1.80} \\ %
\bottomrule
\end{tabular}
\vspace{-0.1cm}
\caption{
\textbf{Multi-agent benchmark} on the AV2 test set.
For all metrics lower values indicate better performance, table sorted in descending order by avgBrierMinFDE$_6$.
For all methods we report results without model ensembling.
}
\vspace{-0.2cm}
\label{tab:res_av2_ma_test}
\end{table*}

\begin{table}[t]
\footnotesize
\centering
\setlength{\tabcolsep}{2pt}
\begin{tabular}{lc|ccccc}
\toprule
Method & Streaming & MR$_6$ & mADE$_6$ & mFDE$_6$ & \textbf{b-mFDE}$_\mathbf{6}$ \\ 
\midrule
SIMPL~\cite{zhang2024simpl}             & \xmark & 0.19 & 0.72 & 1.43 & 2.05 \\
HPTR~\cite{zhang2023hptr}               & \xmark & 0.19 & 0.73 & 1.43 & 2.03 \\
Forecast-MAE~\cite{cheng2023forecast}   & \xmark & 0.17 & 0.71 & 1.39 & 2.03 \\
MTR~\cite{shi2022motion}                & \xmark & 0.15 & 0.73 & 1.44 & 1.98 \\
EMP-D~\cite{prutsch24efficient}         & \xmark & 0.17 & 0.71 & 1.37 & 1.98 \\ 
GANet~\cite{wang2023ganet}              & \xmark & 0.17 & 0.72 & 1.34 & 1.96 \\
ProIn~\cite{dong2024proin}              & \xmark & 0.18 & 0.73 & 1.35 & 1.93 \\
QCNet~\cite{zhou2023query}              & \xmark & 0.16 & 0.65 & 1.29 & 1.91 \\
RealMotion~\cite{song2024realmotion}    & \cmark & 0.15 & 0.66 & 1.24 & 1.89 \\
Tamba~\cite{huang2025trajectory}        & \xmark & 0.17 & 0.64 & 1.24 & 1.89 \\
HAMF~\cite{mei2025hamf}                 & \xmark & 0.14 & 0.64 & 1.23 & 1.89 \\ %
ProphNet~\cite{wang2023prophnet}        & \xmark & 0.18 & 0.68 & 1.33 & 1.88 \\
MTR++~\cite{shi2024mtr++}               & \xmark & 0.14 & 0.71 & 1.37 & 1.88 \\
DyMap~\cite{fan2025bidirectional}       & \xmark & - & 0.71 & 1.29 & 1.87 \\
ModeSeq~\cite{zhou2025modeseq}          & \xmark & \sval{0.14} & \sval{0.63} & 1.26 & 1.87 \\
SmartRefine~\cite{zhou2024smartrefine}  & \xmark & 0.15 & \sval{0.63} & 1.23 & 1.86 \\
MixForecast~\cite{wan2025multi}         & \xmark & \sval{0.14}& 0.64 & 1.22 & 1.86 \\
DeMo~\cite{zhang2024demo}               & \cmark & \bval{0.13} & \bval{0.61} & \bval{1.17} & \sval{1.84} \\
\rowcolor{rcol}\mn~(Ours)               & \cmark & \sval{0.14} & 0.64 & \sval{1.19} & \bval{1.83} \\ %
\bottomrule
\end{tabular}
\vspace{-0.1cm}
\caption{
\textbf{Single-agent benchmark} on the AV2 test set.
For all metrics lower values indicate better, table sorted in descending order by brier-minFDE$_{6}$.
We report results for published works on the official leaderboard without model ensembling.
}
\label{tab:res_av2_sa_test}
\end{table}

\subsection{Evolving Scene Evaluation}
\Cref{tab:res_av2_evolving} provides a detailed evaluation for streaming trajectory prediction on the AV2 validation set.
The results highlight that our model excels across different context lengths (\cl) and prediction timesteps (\pt), providing accurate prediction on varying number of streaming steps.
Despite achieving strong results on the standard benchmark, related work on streaming trajectory prediction~\cite{song2024realmotion, zhang2024demo} does not generalize across different setups: Their standard setup utilizes three prediction passes covering an overall context length of \cl\,=\,5\,s.
When run using different numbers of streaming steps, as in the continuous setting of a deployed automated vehicle, their forecasting performance degrades notably.
These experiments demonstrate that our model design and training scheme lead to robust and accurate results, despite highly varying context lengths.

\subsection{Argoverse 2 Results}
\Cref{tab:res_av2_ma_test} presents results on the AV2 multi-agent test set, comparing our method to the current state-of-the-art.
Our \mn~achieves strong performance, substantially outperforming prior streaming- and snapshot-based methods.
This demonstrates that our design---accurate marginal predictions combined with a lightweight fusion module for joint reasoning---is effective for multi-agent trajectory forecasting.
Additionally, \Cref{tab:res_av2_sa_test} demonstrates our approach on the AV2 single-agent test set.
Here again, our method achieves highly accurate results.
In particular, despite being able to handle flexible context lengths, our trajectory predictions are within only a few centimeters of the state-of-the-art.
However, the correct predictions are more confident as demonstrated by our favorable Brier score.

\subsection{Ablation Study}

\begin{table}[t]
\centering
\setlength{\tabcolsep}{3pt}
\footnotesize{
\begin{tabular}{ccccc|ccc}
\toprule
Dataset & \cl & TCF & IA & DT & mADE$_6$ & mFDE$_6$ & \textbf{b-mFDE}$_\mathbf{6}$ \\ 
\midrule
     & \multirow{4}{*}{5\,s} & \xmark & \xmark & \xmark & 0.74 & 1.28 & 1.91 \\
     & & \cmark & \xmark & \xmark & 0.64 & 1.22 & 1.84 \\ %
     & & \cmark & \cmark & \xmark & \bval{0.63} & \bval{1.19} & \bval{1.81} \\ %
AV2  & & \cc \cmark & \cc \cmark & \cc \cmark & \cc \cc 0.64 & \cc 1.20 & \cc 1.82 \\ %
\cmidrule(l){2-8}
Val  & \multirow{4}{*}{1\,s} & \xmark & \xmark & \xmark & 1.18 & 2.57 & 3.28 \\
     & & \cmark & \xmark & \xmark & 1.09 & 2.49 & 3.18 \\ %
     & & \cmark & \cmark & \xmark & 1.11 & 2.55 & 3.25 \\ %
     & & \cc \cmark & \cc \cmark & \cc \cmark & \cc \cc \bval{0.76} & \cc \bval{1.48} & \cc \bval{2.13} \\ %
\midrule
    & & \cmark & \xmark & \xmark & 0.62 & 0.96 & 1.58 \\ %
    & 2s & \cmark & \cmark & \xmark & \bval{0.59} & \bval{0.91} & \bval{1.52} \\ %
AV1 & & \cc \cmark & \cc \cmark & \cc  \cmark & \cc \bval{0.59} & \cc \bval{0.91} & \cc 1.53 \\ %
\cmidrule(l){2-8}
Val & & \cmark & \xmark & \xmark & 0.75 & 1.22 & 1.86 \\ %
    & 0.5s & \cmark & \cmark & \xmark & 0.71 & 1.13 & 1.78    \\ %
    & & \cc \cmark & \cc \cmark & \cc \cmark & \cc \bval{0.64} & \cc \bval{1.00} & \cc \bval{1.61} \\ %
\bottomrule
\end{tabular}
}
\vspace{-0.1cm}
\caption{
\textbf{Ablation study} on the impact of our target-centric features~(TCF), Dual Training~(DT) and \mbox{Instance-Aware~(IA)} Context Streaming using varying context lengths (\cl).
For each dataset, we compare full-history and partial-history settings.
All predictions adhere to the standard evaluation protocol: \pt$=2\,s$, \fh$=3\,s$ for AV1 and \pt$=5\,s$, \fh$=6\,s$ for AV2.
}
\vspace{-0.4cm}
\label{tab:abl}
\end{table}

We conduct an ablation study in \Cref{tab:abl}, where the baseline performance is obtained with standard information streaming~\cite{song2024realmotion}.
Incorporating our target-centric features, refined via context from previously predicted trajectory endpoints, improves performance in both context settings.
Our instance-aware streaming mechanism is especially favorable for longer historical contexts.
However, using only the instance-aware streaming mechanism biases the model toward long-term dependencies, resulting in reduced performance for short inputs.
Adding also our dual training scheme improves the performance for short contexts notably (\eg mFDE$_6$ -1\,m on AV2), as it encourages the model to learn a more robust temporal representation and to generalize across varying context lengths.
This added robustness comes at a negligible accuracy decrease (\eg mFDE$_6$ +0.01\,m on AV2) when evaluated with long input contexts.

\subsection{nuScenes Results}
The short context length of nuScenes (2 sec at only 2\,Hz) makes it a particularly challenging dataset, which is commonly not used for testing streaming-based motion forecasting at all.
However, our \mn~is designed to handle varying context lengths and therefore performs well even under such constrained contexts, as shown in \cref{tab:res_nuscenes}.

\begin{table}[t]
\centering
\setlength{\tabcolsep}{6pt}
\footnotesize{
\begin{tabular}{lc|ccc}
\toprule
Method & Streaming & mFDE$_1$ & MR$_5$ & mADE$_5$ \\ 
\midrule
THOMAS~\cite{gilles2022thomas} & \xmark & 6.71 & 0.55 & 1.33 \\
PGP~\cite{deo2021multimodal} & \xmark & 7.17 & 0.52 & 1.27 \\
DeMo~\cite{zhang2024demo} & \xmark & 6.60 & 0.43 & 1.22 \\
MacFormer~\cite{feng2023macformer} & \xmark & 7.50 & 0.57 & 1.21 \\
LAFormer~\cite{liu2024laformer} & \xmark & 6.95 & 0.48 & 1.19 \\
Q-EANet~\cite{chen2024q} & \xmark & 6.77 & 0.48 & 1.18 \\
FRM~\cite{park2023leveraging} & \xmark & \sval{6.59} & 0.48 & 1.18 \\
CASPFormer~\cite{yadav2024caspformer} & \xmark & 6.70 & 0.48 & 1.15 \\
LMFormer~\cite{yadav2025lmformer} & \xmark & 6.85 & 0.47 & 1.14 \\
Goal-LBP~\cite{yao2023goal} & \xmark & 9.20 & \sval{0.32} & \bval{1.02} \\
\rowcolor{rcol} \mn~(Ours) & \cmark & \bval{6.02} & \bval{0.28} & \sval{1.13} \\ %
\bottomrule
\end{tabular}
}
\vspace{-0.1cm}
\caption{
Results on the \textbf{nuScenes prediction challenge} test set.
Following the evaluation protocol, we set the model output to \mbox{$K=5$}.
DeMo~\cite{zhang2024demo} reports nuScenes results without streaming.
}
\label{tab:res_nuscenes}
\vspace{-0.15cm}
\end{table}

\subsection{Comparison to Varying-Length Methods}
\Cref{tab:res_obs_len} presents a comparison with recent methods that explicitly target variable observation lengths: FLN~\cite{xu2024adapting}, POP~\cite{wang2024improving}, and TSD~\cite{zhu2025target}.
Our approach achieves favorable results across all datasets and input durations, demonstrating better adaptability to diverse temporal contexts.
These results highlight that explicitly modeling heterogeneous observation lengths through a progressively extending streaming process not only aligns with real-world dynamics but also delivers significant and consistent performance improvements over existing approaches.

\subsection{Latency Comparison}
Our approach achieves low latency by employing an efficient transformer-based backbone without introducing resource-intensive components, such as refinement steps during decoding.
Since available codebases restrict comparison to the AV2 single-agent setting, we report results accordingly.
On a single NVIDIA RTX 3090 GPU, our method achieves an online latency of 33\,ms for predicting batches of 16 agents, compared to 35\,ms for DeMo~\cite{zhang2024demo} and 44\,ms for RealMotion~\cite{song2024realmotion}; 35\,ms (\textit{vs.} 49\,ms and 79\,ms) for 32 agents; and 57\,ms (\textit{vs.} 88\,ms and 140\,ms) for 64 agents.
In the AV2 multi-agent setting, our approach achieves an average online latency of 30\,ms per scenario on the same GPU.

\begin{table}[t]
\centering
\setlength{\tabcolsep}{4pt}
\footnotesize{
\begin{tabular}{ccc|ccc}
\toprule
Dataset & Input Steps & \cl & Method & mADE$_6$ & mFDE$_6$ \\ 
\midrule
\multirow{11}{*}{\rotatebox[origin=c]{90}{Argoverse 1}} & \multirow{3}{*}{5} & \multirow{3}{*}{0.5\,s} & POP-H~\cite{wang2024improving} & 0.74 & 1.10 \\
&&& TSD-H~\cite{zhu2025target} & 0.72 & 1.04 \\
&&& \cc\mn~(Ours)                       &\cc \bval{0.61} &\cc \bval{0.96} \\ \cmidrule(l){2-6} %
& \multirow{4}{*}{10} & \multirow{4}{*}{1\,s} & HiVT-64-FLN~\cite{xu2024adapting} & 0.81    & 1.25 \\
&&& POP-H~\cite{wang2024improving} & 0.70 & 1.06 \\
&&& TSD-H~\cite{zhu2025target} & 0.67 & 0.98 \\
&&& \cc\mn~(Ours)                       &\cc \bval{0.61} &\cc \bval{0.96} \\ \cmidrule(l){2-6}  %
& \multirow{4}{*}{20} & \multirow{4}{*}{2\,s} & HiVT-64-FLN~\cite{xu2024adapting} & 0.72    & 1.08 \\
&&& POP-H~\cite{wang2024improving} & 0.69 & 1.04 \\
&&& TSD-H~\cite{zhu2025target} & 0.65 & 0.94 \\
&&& \cc\mn~(Ours)                       &\cc \bval{0.59} &\cc \bval{0.91} \\ %
\midrule
\multirow{6}{*}{\rotatebox[origin=c]{90}{Argoverse 2}} & \multirow{2}{*}{10} & \multirow{2}{*}{1\,s} & POP-Q~\cite{wang2024improving} & 2.24 & 4.16 \\
  & & &\cc\mn~(Ours)                          &\cc \bval{0.76} &\cc \bval{1.48} \\ \cmidrule(l){2-6}
  & \multirow{2}{*}{30} & \multirow{2}{*}{3\,s} & POP-Q~\cite{wang2024improving} & 0.92 & 1.70 \\
  & & &\cc\mn~(Ours)                          &\cc \bval{0.73} &\cc \bval{1.44} \\ \cmidrule(l){2-6}
  & \multirow{2}{*}{50} & \multirow{2}{*}{5\,s} & POP-Q~\cite{wang2024improving} & 0.79 & 1.45 \\
  & & &\cc\mn~(Ours)                          &\cc \bval{0.64} & \cc \bval{1.20} \\
\midrule
Dataset & Input Steps & \cl & Method & mADE$_5$ & mFDE$_5$ \\ 
\midrule
\multirow{6}{*}{\rotatebox[origin=c]{90}{nuScenes}} & \multirow{2}{*}{2} & \multirow{2}{*}{0.5\,s} & AFormer-FLN~\cite{xu2024adapting}    & 1.92    & 3.91 \\
  & & &\cc\mn~(Ours)                         & \bvc{1.18} & \bvc{2.01} \\  \cmidrule(l){2-6}
& \multirow{2}{*}{3} & \multirow{2}{*}{1.5\,s} & AFormer-FLN~\cite{xu2024adapting}    & 1.88    & 3.89 \\
& & &\cc\mn~(Ours)                         & \bvc{1.15} & \bvc{2.00} \\  \cmidrule(l){2-6}
& \multirow{2}{*}{4} & \multirow{2}{*}{2\,s} & AFormer-FLN~\cite{xu2024adapting}    & 1.83    & 3.78 \\
 & & &\cc\mn~(Ours)                         & \bvc{1.13} & \bvc{1.92} \\
\bottomrule
\end{tabular}
}
\vspace{-0.1cm}
\caption{
Comparison to the state-of-the-art \textbf{varying-length models}~\cite{xu2024adapting, wang2024improving, zhu2025target}, following their evaluation protocol.
}
\vspace{-0.5cm}
\label{tab:res_obs_len}
\end{table}

\section{Conclusion}
\label{sec:conclusion}
We presented a streaming-based motion forecasting approach that enables continuous and efficient prediction in dynamically evolving scenes.
Our \mn~achieves competitive accuracies in the single-agent settings and sets a new state-of-the-art for streaming multi-agent prediction.
Extensive experiments on AV2, AV1, and nuScenes demonstrate strong robustness across diverse temporal contexts and scene complexities.
By explicitly modeling robust context propagation, our work advances real-time trajectory prediction and highlights the importance of streaming processing.

{\small
\vspace{6pt}
\nbf{Acknowledgments}
This work was partially funded by Addsafety (923936), a COMET Project funded by BMIMI, BMWET and the co-financing federal province of Styria.
The COMET programme is managed by the Austrian Research Promotion Agency (FFG).

}

{\small
\bibliographystyle{ieeenat_fullname}
\bibliography{_sections/_references}
}

\ifarxiv \clearpage \appendix 
\maketitlesupplementary
In this supplementary, we first present additional insights to highlight the effectiveness and robustness of our approach, which were omitted from the original paper due to the page limit (\Cref{sec:ae}).
To support reproducibility, we provide all implementation details of our architecture in \Cref{sec:id}, including training hyperparameters and details on the multi-agent extension.
We further detail the evaluation setup (\cref{sec:ed}) and provide output visualizations (\cref{sec:v}).
Finally, we provide an overview of the code framework (\cref{sec:ie}), which is provided at \mbox{\url{https://github.com/a-pru/sharp}}.

\section{Additional Insights}
\label{sec:ae}
\subsection{Observation Length Study}
\Cref{tab:ols} provides an ablation study for predictions of our \mn~across different context lengths \cl.
The results indicate that our approach provides better prediction performance with increasing input context length, highlighting the advantage of a model that effectively exploits extended historical information -- a key motivation for designing \mn.

\subsection{Robustness to Noise}
\Cref{tab:noise_ia} evaluates our instance-aware context streaming under noisy instance masks.
We apply two types of perturbations: (1)~We randomize mask values, introducing false-positive associations between unrelated instances and removing correct correspondences; and (2)~We remove existing associations, simulating ID switches during tracking.
The results indicate that our instance-aware context streaming works well, even under perturbed instance masks.
This means that our design can effectively leverage explicit instance correspondence information when available, while still falling back on implicit context fusion driven by global agent positions~\cite{song2024realmotion}.

Additionally, \Cref{tab:noise_tcf} evaluates the effect of perturbing prediction endpoints during streaming.
This experiment assesses the robustness of our target-centric features to errors in past predictions (\ie, error propagation).
We add Gaussian~($\mathcal{N}$) or uniform~($\mathcal{U}$) noise to the endpoint locations before using them to aggregate the target-centric features.
While these perturbations cause a slight degradation, our approach effectively uses both agent- and target-centric features, enabling accurate predictions even when previous outputs are imperfect.

\renewcommand{\pl}[0]{\multicolumn{3}{c}{\cellcolor{white}\phantom{a}}}
\renewcommand{\win}[1]{\multirow{2}{*}{\includegraphics[page=#1, width=0.5\linewidth]{figures/raw_figures/streaming_windows/supp_windows.pdf}}}

\begin{table}[t]
\centering
\setlength{\tabcolsep}{4pt}
\footnotesize{
\begin{tabular}{c|ccc}
\toprule
Streaming Setup & mADE$_6$ & mFDE$_6$ & \textbf{b-mFDE}$_\mathbf{6}$ \\ 
\midrule
\win{5} & \multirow{2}{*}{0.76} &  \multirow{2}{*}{1.48} &  \multirow{2}{*}{2.13} \\ 
&& \\
\win{6} & \multirow{2}{*}{0.73} &  \multirow{2}{*}{1.44} & \multirow{2}{*}{2.09} \\  
&&  \\ 
\win{7} & \multirow{2}{*}{\bval{0.64}} &  \multirow{2}{*}{\bval{1.20}} & \multirow{2}{*}{\bval{1.82}} \\ 
&&  \\ 
\midrule
\win{8} & \multirow{2}{*}{0.55} &  \multirow{2}{*}{1.11} &  \multirow{2}{*}{1.77} \\ 
&& \\
\win{9} & \multirow{2}{*}{0.49} &  \multirow{2}{*}{0.89} & \multirow{2}{*}{1.55} \\ 
&&  \\ 
\win{10} & \multirow{2}{*}{\bval{0.39}} &  \multirow{2}{*}{\bval{0.70}} & \multirow{2}{*}{\bval{1.31}} \\ 
&&  \\ 
\midrule
\win{11} & \multirow{2}{*}{0.40} &  \multirow{2}{*}{0.72} &  \multirow{2}{*}{1.38} \\ 
&& \\
\win{12} & \multirow{2}{*}{0.41} &  \multirow{2}{*}{0.75} &  \multirow{2}{*}{1.42} \\ 
&& \\
\win{13} & \multirow{2}{*}{0.36} &  \multirow{2}{*}{0.59} & \multirow{2}{*}{1.23} \\ 
&&  \\ 
\win{14} & \multirow{2}{*}{\bval{0.28}} &  \multirow{2}{*}{\bval{0.48}} & \multirow{2}{*}{\bval{1.08}} \\ 
&&  \\ 
\bottomrule
\end{tabular}
}
\vspace{-0.1cm}
\caption{
Evaluation of \mn~on the AV2 validation set performed \textbf{across different context lengths} \cl.
Within each group, the prediction time step \pt~and forecasting horizon \fh~are held constant, while the observation length provided to the model is varied.
We evaluate predictions at the standard \pt$=5\,s$, as well as at \pt$=7\,s$ and \pt$=8\,s$, to study the impact of longer observation lengths on forecasting performance.
\vspace{-0.1cm}
}
\label{tab:ols}
\end{table}

\begin{table}[t]
\centering
\setlength{\tabcolsep}{4pt}
\footnotesize{
\begin{tabular}{cc|ccccccc}
\toprule
Noise Type & Noise Ratio & mADE$_6$ & mFDE$_6$ & b-mFDE$_6$ \\ 
\midrule
\multirow{3}{*}{Randomized} & 80\% & 0.65 & 1.23 & 1.85 \\ %
& 40\% & 0.65 & 1.22 & 1.84 \\ %
& 20\% &  0.64 & 1.21 & 1.83 \\ %
\midrule
\multirow{3}{*}{Set to Zero (no match)} & 80\% & 0.65 & 1.22 & 1.84 \\ %
& 40\% & 0.64 & 1.20 & 1.83 \\ %
& 20\% & 0.64 & 1.20 & 1.82 \\ %
\midrule
\multicolumn{2}{c|}{Clean Data} & 0.63 & 1.19 & 1.81 \\ %
\bottomrule
\end{tabular}
}
\vspace{-0.1cm}
\caption{
We evaluate the \textbf{robustness of our instance-aware context streamer under perturbed instance masks}.
In the first group, we partially randomize the masks, introducing incorrect correspondences and removing correct ones, resulting in both false positives and false negatives.
In the second group, we remove associations by partially setting mask values to zero, producing false negatives while leaving other annotations intact.
\vspace{-0.2cm}
}
\label{tab:noise_ia}
\end{table}

\newcommand{\winevl}[1]{\multirow{3}{*}{\includegraphics[page=#1]{figures/raw_figures/streaming_windows/evolving_scenes_table.pdf}}}

\begin{table*}[t]
\footnotesize
\centering
\setlength{\tabcolsep}{6pt}
\begin{tabular}{c|cccc}
\toprule
History / Prediction Time Step / Future Horizon& avgMinADE$_6$ & avgMinFDE$_6$ & actorMR$_6$ & \textbf{avgBrierMinFDE}$_\mathbf{6}$ \\ 
\midrule
\winevl{2} & \\
& 0.68 & 1.43 & 0.19 & 2.07 \\ %
&& \\
\midrule
\winevl{3} & \\
& 0.60 & 1.26 & 0.17 & 1.91 \\
&& \\
\midrule
\winevl{4} & \\
&  0.54 & 1.17 & 0.15 & 1.83 \\
&& \\
\midrule
\winevl{5} & \\
& 0.43 & 0.88 & 0.11 & 1.52 \\
&& \\
\midrule
\winevl{6} & \\
& 0.32 & 0.64 & 0.06 & 1.27 \\
&& \\
\midrule
\winevl{7} & \\
& 0.23 & 0.43 & 0.03 & 1.06 \\
&& \\
\midrule
\winevl{8} & \\
& 0.55 & 1.16 & 0.15 & 1.80 \\
&& \\
\midrule
\winevl{9} & \\
& 0.56 & 1.21 & 0.16 & 1.86 \\
&& \\
\midrule
\winevl{10} & \\
& 0.46 & 0.97 & 0.12 & 1.62 \\
&& \\
\midrule
\winevl{11} & \\
& 0.55 & 1.16 & 0.15 & 1.80 \\ 
&& \\
\bottomrule
\end{tabular}
\vspace{-0.1cm}
\caption{
Evaluation of \mn~in the \textbf{multi-agent setting} on the AV2 validation set at \textbf{varying prediction time steps} \pt~\textbf{and context lengths} \cl.
The evaluation scope is limited to our method due to missing multi-agent implementation details and code bases for related work~\cite{song2024realmotion, zhang2024demo}.
}
\vspace{-0.3cm}
\label{tab:supp_av2_ma_evolving}
\end{table*}

\begin{table}[t]
\centering
\setlength{\tabcolsep}{6pt}
\footnotesize{
\begin{tabular}{c|ccccccc}
\toprule
Endpoint Noise (meter) & mADE$_6$ & mFDE$_6$ & b-mFDE$_6$ \\ 
\midrule
$\mathcal{U}(-5, 5)$ & 0.65 & 1.23 & 1.85 \\
$\mathcal{N}(0, 3)$\phantom{-0}& 0.65 & 1.23 & 1.85 \\  
$\mathcal{U}(-3, 3)$ & 0.64 & 1.21 & 1.83 \\
$\mathcal{N}(0, 1)$\phantom{-0} & 0.61 & 1.20 & 1.82 \\  
$\mathcal{U}(-1, 1)$ & 0.64 & 1.20 & 1.83 \\
\midrule
Clean Data & 0.63 & 1.19 & 1.81 \\ %
\bottomrule
\end{tabular}
}
\vspace{-0.1cm}
\caption{
We evaluate the \textbf{robustness of our target-centric features under perturbations} of the predicted endpoints used for the token selection.
Gaussian~($\mathcal{N}$) and uniform~($\mathcal{U}$) noise are applied to simulate degradation of predictions during streaming.
This setup assesses how our model handles such errors and its ability to recover from inaccurate prior predictions.
}
\vspace{-0.4cm}
\label{tab:noise_tcf}
\end{table}

\subsection{AV2 Multi-Agent Analysis}
\nbf{Evolving Scenes}
Analogous to Table 1 in the main paper (\emph{AV2 single-agent setting}), \Cref{tab:supp_av2_ma_evolving} analyzes the performance of \mn~across varying context lengths \cl~and prediction time steps \pt~in the \emph{AV2 multi-agent setting}.
The results demonstrate that our approach remains effective under evolving multi-agent scene conditions, further highlighting the robustness and adaptability of our method.

\nbf{Scene Consistency}
To validate our approach for the extension to scene-wide joint predictions, we evaluate the benefit of our joint refinement module over a naive combination of single-agent marginals by measuring potential collision rates on the AV2 multi-agent validation set.
We define a potential collision as an event where two predicted actors within the same predicted world are closer than 2\,m to each other at the same time step.
Utilizing our joint refinement module reduces the number of collision events from 5,533 (naive combination of marginals) to 973.
For reference, the ground truth trajectories contain zero collisions.

\subsection{AV1 Validation Set}
\Cref{tab:av1} compares our approach to the state-of-the-art on the AV1~\cite{chang2019argoverse} validation set.
Overall, SHARP achieves top performance, closely matching the results of HPNet~\cite{tang2024hpnet}.

\subsection{Short Histories}
\begin{table}[t]
\centering
\setlength{\tabcolsep}{6pt}
\footnotesize{
\begin{tabular}{cc|ccc}
\toprule
Method & Streaming & MR$_6$ & mADE$_6$ & \textbf{mFDE}$_\mathbf{6}$\\ 
\midrule
mmTransformer~\cite{liu2021multimodal} & \xmark & 0.11 & 0.71 & 1.15 \\
FRM~\cite{park2023leveraging} & \xmark &  - & 0.68 & 0.99 \\
ADAPT~\cite{aydemir2023adapt} & \xmark & 0.08 & 0.67 & 0.95 \\
SIMPL~\cite{zhang2024simpl} & \xmark & 0.08 & 0.66 & 0.95 \\
HiVT~\cite{zhou2022hivt} & \xmark & 0.09 & 0.66 & 0.96 \\
R-Pred~\cite{choi2023r} & \xmark & 0.09 & 0.66 & 0.95 \\
DeMo~\cite{zhang2024demo} & \cmark & \bval{0.07} & 0.59 & 0.90 \\
\rowcolor{rcol} \mn~(Ours) & \cmark & \bval{0.07} & \bval{0.58} & 0.90 \\ 
HPNet~\cite{tang2024hpnet} & \xmark & \bval{0.07} & 0.64 & \bval{0.87} \\
\bottomrule
\end{tabular}
}
\vspace{-0.1cm}
\caption{
Results on the \textbf{Argoverse 1 prediction challenge validation set}.
}
\vspace{-0.2cm}
\label{tab:av1}
\end{table}

\newcommand{\winsh}[1]{\multirow{2}{*}{\includegraphics[page=#1, width=0.42\linewidth]{figures/raw_figures/streaming_windows/supp_windows.pdf}}}

\begin{table}[t]
\centering
\setlength{\tabcolsep}{2pt}
\footnotesize{
\begin{tabular}{cc|ccc}
\toprule
Streaming Setup & Method & mADE$_6$ & mFDE$_6$ & \textbf{b-mFDE}$_\mathbf{6}$ \\ 
\midrule
\winsh{15} & DeMo & 1.02 & 1.86 & 2.54 \\   
& \cc \mn & \cc \bval{0.70} & \cc \bval{1.16} & \cc \bval{1.79} \\
\midrule
\winsh{2}& DeMo & 0.80 & 1.46 & 2.15 \\
& \cc \mn & \cc \bval{0.70} & \cc \bval{1.34} & \cc \bval{1.99} \\ 
\midrule
\winsh{3} & DeMo & 0.99 & 1.97 & 2.67 \\
& \cc \mn & \cc \bval{0.76} & \cc \bval{1.48} & \cc \bval{2.13} \\ 
\midrule
\winsh{4} & DeMo & 0.95 & 1.96 & 2.67 \\   
& \cc \mn & \cc \bval{0.65} & \cc \bval{1.27} & \cc \bval{1.92} \\
\bottomrule
\end{tabular}
}
\vspace{-0.1cm}
\caption{
We compare our \mn~to DeMo~\cite{zhang2024demo} using \textbf{short context lengths \cl~of 1\,s} on the AV2 single-agent validation set.}
\vspace{-0.2cm}
\label{tab:short_histories}
\end{table}

\Cref{tab:short_histories} compares our method to DeMo~\cite{zhang2024demo} using a 1\,s input length \cl~on the AV2 validation set.
Our approach outperforms DeMo across all prediction horizons \pt, highlighting the effectiveness of our dual training and processing scheme.

\subsection{Extended Prediction Horizon}
\begin{table}[t]
\centering
\setlength{\tabcolsep}{4pt}
\footnotesize{
\begin{tabular}{c|ccc}
\toprule
Streaming Setup & mADE$_6$ & mFDE$_6$ & \textbf{b-mFDE}$_\mathbf{6}$ \\ 
\midrule
\win{16} & \multirow{2}{*}{1.37} & \multirow{2}{*}{2.48} & \multirow{2}{*}{3.12} \\ 
&& \\
\win{17} & \multirow{2}{*}{1.12} & \multirow{2}{*}{2.06} & \multirow{2}{*}{2.70} \\ 
&&  \\ 
\win{18} & \multirow{2}{*}{0.95} & \multirow{2}{*}{1.76} & \multirow{2}{*}{2.40} \\ 
&&  \\ 
\win{19} & \multirow{2}{*}{0.79} & \multirow{2}{*}{1.48} & \multirow{2}{*}{2.11} \\ 
&&  \\ 
\win{20} & \multirow{2}{*}{\bval{0.64}} &  \multirow{2}{*}{\bval{1.20}} & \multirow{2}{*}{\bval{1.82}} \\ 
&&  \\ 
\bottomrule
\end{tabular}
}
\vspace{-0.1cm}
\caption{
Evaluation of \mn~on \textbf{extend prediction horizons} using the AV2 single-agent validation set.
Predictions of the agent’s final position at the end of each scenario are made at varying prediction times \pt.
Even the initial predictions over long horizons provide reasonable estimates.
As the available context \cl~increases and the prediction horizon \fh~shortens, the forecasts naturally become more accurate.
\vspace{-0.4cm}
}
\label{tab:supp_var_futu}
\end{table}

Our short-window processing scheme enables training with extended prediction horizons without requiring additional data.
For instance, AV2 provides scenarios of 11\,s duration.
Within this setting, our model supports a 10\,s prediction horizon, compared to the default 6\,s horizon supported by baselines.
\Cref{tab:supp_var_futu} reports experiments evaluating our module on long prediction horizons \fh.

\subsection{Dual Loss Ablation}
\begin{table}[t]
\centering
\setlength{\tabcolsep}{4pt}
\footnotesize{
\begin{tabular}{cc|ccc}
\toprule
\cl & Loss $\mathcal{L}$ Configuration & mADE$_6$ & mFDE$_6$ & \textbf{b-mFDE}$_\mathbf{6}$ \\ 
\midrule
& $ 2 \cdot\mathcal{L}_\text{stream} +1 \cdot \mathcal{L}_\text{chunk}$ & 0.66 & 1.23 & 1.85  \\ %
5\,s & $1 \cdot\mathcal{L}_\text{stream} +2 \cdot \mathcal{L}_\text{chunk}$ & 0.66 & 1.24 & 1.87 \\ %
& $1 \cdot\mathcal{L}_\text{stream} +1 \cdot \mathcal{L}_\text{chunk}$ & 0.66 & 1.24 & 1.86 \\ %
\midrule
&$2 \cdot\mathcal{L}_\text{stream} +1 \cdot \mathcal{L}_\text{chunk}$ & 0.97 & 2.14 & 2.82  \\ %
1\,s & $1 \cdot\mathcal{L}_\text{stream} + 2 \cdot \mathcal{L}_\text{chunk}$ & 0.87 & 1.84 & 2.52 \\ %
 & $1 \cdot\mathcal{L}_\text{stream} +1 \cdot \mathcal{L}_\text{chunk}$ & 0.94 & 2.03 & 2.70 \\ %
\bottomrule
\end{tabular}
}
\vspace{-0.1cm}
\caption{
We evaluate training \mn~using a \textbf{weighted dual loss formulation}, where the weights allow us to emphasize either long-term or short-term context performance.
In our final model, we omit these weights, resulting in equal contributions from both loss terms.
Note that this analysis is based on a preliminary experiment conducted with a reduced training schedule.
}
\vspace{-0.2cm}
\label{tab:abl_dual}
\end{table}

\Cref{tab:abl_dual} presents results for training our model using weighting parameters to combine the two losses, $\mathcal{L}_{\text{stream}}$ and $\mathcal{L}_{\text{chunk}}$, in our dual-training formulation.
The experiment shows that adjusting the loss weights enables the model to prioritize either short-term context or longer prediction horizons, and further highlights the effectiveness of our loss design.
To obtain the best overall trade-off, we use equal weighting between the two losses in our final implementation.

\subsection{Context Streaming Ablation Study}
\begin{table}[t]
    \centering
    \setlength{\tabcolsep}{1.3pt}
    \footnotesize{
    \begin{tabular}{cc|ccc}
        \toprule
        Streaming Setup & $f_\text{IA}$ & mADE$_6$ & mFDE$_6$ & \textbf{b-mFDE}$_\mathbf{6}$ \\ 
        \midrule
        \win{21} & \xmark & 0.54 & 0.99 & 1.63 \\ 
         & \cmark & \textbf{0.51} & \textbf{0.94} & \textbf{1.56} \\ 
        \midrule
        \win{10} & \xmark & 0.47 & 0.84 & 1.50 \\ 
         & \cmark & \textbf{0.39} & \textbf{0.70} & \textbf{1.31} \\ 
        \midrule
        \win{14} & \xmark & 0.39 & 0.66 & 1.33 \\  %
         & \cmark & \textbf{0.28} & \textbf{0.48} & \textbf{1.08} \\ 
        \bottomrule
    \end{tabular}
    }
    \vspace{-0.2cm}
    \caption{
    Ablation study on AV2 for long input horizons using our \textbf{instance-aware context streaming} $f_\text{IA}$ module compared to the standard context relay streaming module proposed by~\cite{song2024realmotion}.
    \vspace{-0.4cm}
    }
\label{tab:abl_ia}
\end{table}

\Cref{tab:abl_ia}~compares our instance-aware context streamer against the implicit approach of~\cite{song2024realmotion} across extended observation lengths \cl.
The results demonstrate that explicit instance modeling provides a significant advantage, particularly as the observation duration increases.

\begin{table}[t]
\centering
\setlength{\tabcolsep}{4pt}
\footnotesize{
\begin{tabular}{ccc|ccc}
\toprule
$T_h$ & Train Pass at \pt & IA & mADE$_6$ & mFDE$_6$ & \textbf{b-mFDE}$_\mathbf{6}$ \\ 
\midrule
3\,s & \{3, 4, 5\}\,s       & \xmark & 0.66 & 1.23 & 1.84 \\ %
\midrule
2\,s & 2 .. 5\,s       & \xmark & 0.66 & 1.24 & 1.85 \\ %
1\,s & 1 .. 5\,s       & \xmark & 0.64 & 1.22 & 1.85 \\ %
1\,s & 1 .. 8\,s       & \xmark & 0.65 & 1.22 & 1.84 \\ %
\midrule
2\,s & 2 .. 5\,s       & \cmark & 0.66 & 1.22 & 1.83 \\ %
1\,s & 1 .. 5\,s       & \cmark & 0.65 & 1.21 & 1.83 \\ %
1\,s & 1 .. 8\,s       & \cmark & \bval{0.63} & \bval{1.19} & \bval{1.81} \\ %
\bottomrule
\end{tabular}
}
\vspace{-0.1cm}
\caption{
Ablation study on \textbf{training} our approach with varying input window sizes ($T_h$) and different numbers of back-propagation passes per scenario, with and without our instance-aware context streaming~(IA).
The displacement during consecutive train passes is 1\,s for all experiments.
The first row reports training results for our \mn~architecture using the baseline streaming setup of~\cite{song2024realmotion}.
Reducing the input window size $T_h$ enables performing more back-propagation passes over each scenario.
However, existing streaming mechanisms fail to fully exploit this potential, which is effectively unlocked by our instance-aware context streamer.
All evaluations follow the standard protocol on the AV2 validation set.
Experiments are conducted without dual training to neglect confounding influences and isolate the effect of instance-aware context streaming.
}
\vspace{-0.2cm}
\label{tab:abl_steps}
\end{table}

\subsection{Model Training Ablation Study}
\Cref{tab:abl_steps} analyzes the impact of our instance-aware context streaming across different model setups.
All evaluations follow the standard AV2 protocol, using \cl\,=5\,\,s of historical context and predictions at \pt\,=\,5\,s for 6\,s into the future.
We vary the input observation window size ($T_h$) and the number of backpropagation passes during training.
The first row corresponds to the default streaming setup of~\cite{song2024realmotion}.
Reducing the input window size allows for creating more (non-overlapping) train samples per dataset scenario.
However, the baseline architecture degrades due to limited temporal information per pass and restricted inter-frame propagation through the implicit agent-relation modeling.
In contrast, our instance-aware context streaming significantly improves performance by preserving temporal coherence across streaming steps.
The best results are achieved when training with eight model passes, which increases sample diversity and enables more effective learning of the streaming mechanism.

\subsection{Latency Analysis}
\begin{table*}[t]
\footnotesize
\centering
\begin{tabular}{lc|ccccc|c|c}
\toprule
Method & Streaming & $B=1$ & $B=16$ & $B=32$ & $B=64$ & $B=128$ & Model Parameters & AV2 Test Set brier-minFDE$_{6}$\\
\midrule
QCNet~\cite{zhou2023query}              & \xmark & 141\,ms & 303\,ms & 565\,ms & 1092\,ms & - & 7.7M & 1.91 \\
RealMotion~\cite{song2024realmotion}    & \cmark & \bval{19\,ms} & 44\,ms & \phantom{0}79\,ms & 140\,ms & 273\,ms & \bval{2.9M} & 1.89 \\
DeMo~\cite{zhang2024demo}               & \cmark & \phantom{0}25\,ms & 35\,ms & \phantom{0}49\,ms & \phantom{00}88\,ms & 180\,ms  & 5.9M & 1.84 \\
\rowcolor{rcol}\mn~(Ours)               & \cmark & \phantom{0}21\,ms & \bval{33\,ms} & \phantom{0}\bval{35\,ms} & \phantom{00}\bval{57\,ms} & \bval{105\,ms} & 4.6M & \bval{1.83} \\
\bottomrule
\end{tabular}
\vspace{-0.1cm}
\caption{
\textbf{Latency analysis} for batches of size $B$ when performing \emph{single-agent prediction}, evaluated on the Argoverse 2 validation set with an NVIDIA RTX 3090 GPU.
Larger batch sizes $B$ correspond to more practically-relevant latency estimates, since real-world deployments must process multiple surrounding traffic agents, \ie $B>1$, simultaneously.
For QCNet we run out of GPU memory when testing $B=128$.
}
\vspace{-0.2cm}
\label{tab:latency}
\end{table*}

We provide a detailed latency analysis of both streaming and standard methods on a single NVIDIA RTX 3090 GPU in \Cref{tab:latency}.
The results are reported for the \emph{single-agent} setting, where each inference step produces predictions for a single agent.
In real-world scenarios, however, we are interested in forecasting for multiple surrounding traffic agents.
Thus, practically relevant latencies correspond to larger batch sizes $B$, since $ B$ reflects the number of other agents for which predictions are produced.
On the AV2 single-agent dataset, our approach requires under 10 GB of memory for inference with a batch size of 32 ($<$5 GB for batch size 16 and $<$15 GB for batch size 128), further highlighting the resource efficiency of our model.

We use the official implementations of QCNet\footnote{\url{https://github.com/ZikangZhou/QCNet}}~\cite{zhou2023query},
RealMotion\footnote{\url{https://github.com/fudan-zvg/RealMotion}\label{fn:remo}}~\cite{song2024realmotion}, and
DeMo\footnote{\url{https://github.com/fudan-zvg/DeMo}}~\cite{zhang2024demo}. 
To ensure a fair comparison across different code bases, we measure only the model forward-pass latency to eliminate the influence of data preprocessing and loading.

\section{Implementation Details}
\label{sec:id}
In the following, we outline all model details, including module designs, hyperparameters, and training setup.
For all experiments, we use a model feature dimension of $D=128$.
\Cref{tab:supp_setup} presents an overview of parameters used for loading agent and map data from the datasets (AV1~\cite{chang2019argoverse}, nuScenes~\cite{caesar2020nuscenes}, and AV2~\cite{wilson2021argoverse}).
We also add a valid flag to the agent and lane features to pad lanes, which extend outside the region of interest.
This leads to an agent feature dimension $D_a=5$ for AV2 and $D_a=3$ for AV1 and nuScenes.
Across all datasets, the lane feature dimension is $D_l=3$ ($xy$ coordinates of the center line and padding flag).

\begin{table*}[t]
\centering
\setlength{\tabcolsep}{3pt}
\footnotesize{
\begin{tabular}{c||c|c|c|c|c|c|c}
\toprule
\multirow{2}{*}{Dataset}        & RoI           & \multicolumn{2}{c|}{Agents} & \multicolumn{3}{c|}{Lanes} & Input Window \\
 & Radius & Features & Types & Features & Types & $P_l$ & Length $T_h$ \\ 
\midrule
AV1~\cite{chang2019argoverse} & 150\,m & $xy$-pos & Vehicle & $xy$-pos & Standard Lane & 10 & 0.5\,s \\
AV2~\cite{wilson2021argoverse} & 150\,m & $xy$-pos, $xy$-vel & Veh./Pedest./Cycl./Others & $xy$-pos & Veh./Bike/Bus & 20 & 1\,s \\
nuScenes~\cite{caesar2020nuscenes} & 100\,m & $xy$-pos & Car/Pedest./Cycl./Other Veh./Motorcycl./Others &$xy$-pos & Standard/Intersection & 20 & 0.5\,s \\
\bottomrule
\end{tabular}
}
\vspace{-0.1cm}
\caption{
\textbf{Data preprocessing and sampling parameters} for loading data from Argoverse~1~(AV1), Argoverse~2~(AV2), and nuScenes.
\textit{RoI radius} denotes the region of interest radius used to sample surrounding agent and map data.
}
\vspace{-0.2cm}
\label{tab:supp_setup}
\end{table*}

\subsection{Streaming Processing}
\begin{figure}[tp]
    \centering
        \includegraphics[trim={0cm, 0cm, 0cm, 0cm}, clip, width=0.95\linewidth]{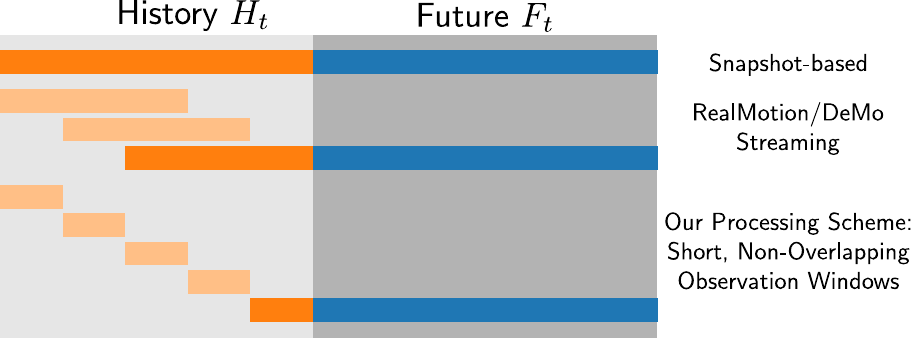} 
    \vspace{-0.2cm}
    \caption{
    Comparison of motion forecasting processing schemes: standard snapshot-based forecasting, the streaming-based forecasting used by RealMotion~\cite{song2024realmotion} and Demo~\cite{zhang2024demo} (long, overlapping, observation windows) and our proposed approach (compact, subsequent, observation windows).
    }
    \label{fig:supp_timeline}
    \vspace{-0.45cm}
\end{figure}

\Cref{fig:supp_timeline} compares our proposed streaming processing scheme to the standard snapshot-based evaluation---which uses the whole history available in a benchmark dataset in one model pass---and to the streaming processing proposed by RealMotion~\cite{song2024realmotion}.
Compared to RealMotion, we use a non-overlapping streaming mechanism with significantly shorter observation lengths per model pass.

\subsection{Target-Centric Features}
A key challenge in trajectory prediction is identifying the map regions where detailed features should be extracted.
Besides the immediate surroundings of each agent, as modeled by standard agent-centric features, the regions an agent is likely to move toward in the future are also highly important.
To encode this information, we introduce a set of target-centric auxiliary features, constructed from the predicted endpoints of the previous inference pass $F^{t-1}$. 
For each endpoint, we establish a local coordinate frame with the endpoint as the origin, and we encode all contextual features within its vicinity into a target-centric feature set.
Based on preliminary experiments, we use a 30\,m radius for aggregating agent and map tokens.
For each of the $K$ predicted endpoints, this yields an auxiliary feature tensor $C_\senc^t \in \mathbb{R}^{K \times (N_a+N_l)' \times D}$, where $(N_a+N_l)'$ denotes the maximum number of features among all $K$ endpoint-centric sets; we pad smaller sets to ensure consistent batch processing.
To incorporate these features during decoding, we additionally encode the spatial transformation between the global coordinate frame and each endpoint-centric local frame through a positional embedding.

\subsection{Multi-Head Attention Blocks}
We use a standard configuration~\cite{cheng2023forecast, prutsch24efficient, song2024realmotion} for our multi-head attention blocks used in the encoder and decoder modules:
\begin{itemize}
    \item 8 Attention heads
    \item Dropout rate set to 0.2
    \item Feedforward path expansion factor set to $4\cdot D$
    \item Norm executed before attention operation
    \item GELU~\cite{hendrycks2016gaussian} Activation function
\end{itemize}

\subsection{Encoder Modules}
\nbf{Agent and Lane Encoder}
Initially, we use a single layer to project the input agent features $A\in \mathbb{R}^{N_a \times T_h \times D_a}$ to our model feature dimension $D$, leading to $\mathbb{R}^{N_a \times T_h \times D}$.
Next, we apply self-attention across the historical time steps using four encoder blocks.
To reduce the dimensionality, we apply a max-pooling layer, resulting in $A_\text{enc}\in \mathbb{R}^{N_a \times D}$ as output for $f_A$.
For encoding the lanes $f_L$ we use the standard mini-PointNet-based~\cite{qi2017pointnet} approach which is commonly used by related work~\cite{cheng2023forecast, prutsch24efficient, zhang2024demo, song2024realmotion}.

\nbf{Positional Embeddings}
We model all positional embeddings using the 2D coordinates ($x,y$) and orientation ($\gamma$).
For agents, we use the agent's pose and heading; for lanes, we use the centerline's midpoint and rotation.
To wrap the angle, we encode each pose as $(x, y, \sin \gamma, \cos \gamma)$.
We implement the positional encodings using a shallow multilayer perceptron~(MLP):
\begin{itemize}
    \item Layer 1: $4 \times D$
    \item[] GELU~\cite{hendrycks2016gaussian} activation function
    \item Layer 2: $D \times D$
\end{itemize}

\nbf{Instance-aware Context Streamer}
Our instance-aware context streamer $f_\text{IA}$ follows the design principle proposed by~\cite{song2024realmotion} and applies cross-attention between the current scene $S^t$ and the previously encoded context $S_\senc^{t-1}$.
As agents enter or leave the scene and the map content within the region of interest changes due to ego-motion, the number of tokens $N_c$ generally differs between the two contexts.
To explicitly model instance correspondences across time between these contexts, we introduce an attention mask $M \in \mathbb{R}^{N_c^t \times N_c^{t-1}}$, where $M_{ij}=\theta$ if token at position $i$ in $S^t$ corresponds to the token at position $j$ in $S_\senc^{t-1}$ and $0$ otherwise. 
The mask weight $\theta$ is a learnable parameter and is jointly optimized during training.
After the context streaming, we obtain $S_\text{stream}^t \in \mathbb{R}^{N_c^t \times D}$. 

\nbf{Scene Context Encoders}
After constructing the agent-centric scene context and augmenting it with the previous context, we encode $S_\text{stream}^{t}$ using the agent-centric scene context encoder $f_S$. 
This encoder consists of four attention blocks, where self-attention is applied across all agent and lane tokens to jointly capture their relationships.
The target-centric feature encoder $f_T$ follows the same design principles but uses only two blocks because $ C_t$ has fewer tokens.

\subsection{Decoder}
Our decoder $f_D$ uses separate cross-attention blocks to attend to the agent-centric scene features $S_\senc^t$ and the target-centric context $C_\senc^t$.
The decoder alternates between attending to $S_\senc^t$ and $C_\senc^t$.
If no target-centric context is available (\eg, for newly detected agents), we simply skip the blocks that would attend to $C_\senc^t$.
In our final architecture, we use three blocks per feature type, resulting in a total of six cross-attention blocks.
After updating the mode queries  $\encmodequery \in \mathbb{R}^{K \times D}$, we apply two MLP heads to produce the trajectory forecasts $F$ and their corresponding prediction scores $P$.
The forecast head has the following design: 
\begin{itemize}
    \item Layer 1: $D \times 2\cdot  D$
    \item[] ReLU activation function
    \item Layer 2: $2\cdot D \times 2\cdot T_f$
\end{itemize}
Similarly, the prediction head is given as: 
\begin{itemize}
    \item Layer 1: $D \times 2\cdot  D$
    \item[] ReLU activation function
    \item Layer 2: $2\cdot D \times 1$
\end{itemize}

\subsection{Optimization}
We train our model using a single NVIDIA Quadro RTX 8000 with 48\,GB VRAM using a batch size of 32.
To accommodate the different dataset complexities, training is executed for 80 epochs on AV2, for 60 epochs on AV1 and for 25 epochs on nuScenes.
For all trainings, we use warm-up epochs (13/10/4), where the learning rate is linearly increased to $1e-4$ before being decreased to $1e-5$ using a single cosine schedule.
We apply gradient clipping and weight decay regularization.
No augmentations are used, and we use no cross-dataset training.

We use AdamW~\cite{loshchilov2017decoupled} as our optimizer.
Following common practice~\cite{cheng2023forecast}, only the best-matching trajectory hypothesis contributes to the loss via a winner-takes-all strategy.
A Smooth L1 regression loss~\cite{huber1964robust} is applied to fit this selected hypothesis to the ground truth, and a cross-entropy classification loss encourages the model to assign the highest probability to it.
Additionally, we include an auxiliary regression head that predicts a single trajectory for other agents surrounding the focal agent~\cite{cheng2023forecast}.
For this auxiliary task, we apply a two-layer MLP to the agent features in $S_\senc^t$ and supervise it using a Smooth L1 loss.

\subsection{Multi-Agent Extension}
\label{sec:app:multiagent}
\nbf{Global Consistency Module}
We employ a global consistency module to combine the marginal predictions --- represented by the per-agent mode queries after cross-attention, $\encmodequery \in \mathbb{R}^{N_a \times K \times D}$ --- into scene-level, jointly consistent forecasts.
We first select a scene-wide reference frame, choosing either a specific agent or the self-driving vehicle; in our implementation, we simply use the first agent listed in each scenario.
Subsequently, we add a positional embedding modeling the current agent position \wrt the scene global coordinate system.
To capture both intra- and inter-agent relationships, we apply self-attention over all motion modes for each agent (across $K$) and across all agents for each hypothesized future world (across $N_a$), using two blocks for each.
The resulting predictions are organized in $K$ worlds, where each world contains predictions for all agents and represents a collision-free, globally consistent future.
Trajectories within each world are generated using a two-layer MLP following the single-agent design described above.
World-level probability scores are obtained by fusing the agent queries associated with that world and feeding the fused representation into an MLP head, again following the single-agent setup.
To avoid the influence of parked vehicles, we exclude them during the fusion step based on the trajectory head's output.

\nbf{Finetuning on Multi-Agent Data}
We finetune our model for 35 epochs on the multi-agent data using a single NVIDIA RTX PRO 6000 with 96\,GB VRAM.
We set the batch size to 32 scenarios and initialize our model for the marginal predictions with the weights trained on the single-agent dataset.
During finetuning, we combine marginal prediction and joint prediction losses to train the global consistency module while also providing additional guidance to the early encoder layers.
The single-agent losses are identical to those in the single-agent training.
For the multi-agent output, we again employ a winner-takes-all strategy, only optimizing the \emph{world} with the lowest overall displacement.
Next, we compute a regression loss based on the prediction for each agent in this world, as well as a standard cross-entropy loss, so that this world receives the highest probability.

\section{Evaluation Details}
\label{sec:ed}

\subsection{Metrics}
We evaluate our approach using the standard benchmark metrics~\cite{chang2019argoverse, caesar2020nuscenes, wilson2021argoverse}.
Each metric is computed over the top-$k$ highest-scoring trajectory forecasts.
In the single-agent setting, we report the \textbf{miss rate} (MR$_k$), which evaluates if any predicted endpoint lies within a radius $r$ to the ground endpoints; \textbf{average displacement error} (minADE$_k$), which reports the average displacement between the ground truth and averaged across all future time steps; and the \textbf{final displacement error} (minFDE$_k$), which reports the smallest distance between the ground truth endpoint and a predicted endpoint.
The \textbf{brier-minFDE}$_k$ adds a probability penalty $(1-p)^2$ for the minFDE$_k$ based on the score $p$ for the trajectory with the lowest displacement.

For the multi-agent setting, we use the joint extensions of the single-agent metrics~\cite{wilson2021argoverse}: avgMinADE$_k$, avgMinFDE$_k$, actorMR$_k$, and avgBrierMinFDE$_6$.
Here, $k$ worlds are considered, where each world contains one prediction for every agent.
Each metric is computed per world by averaging over all agent errors in this world.

\subsection{Evaluation Protocol for Main \mbox{Paper} Table 1}
\begin{figure}[tp]
    \centering
        \includegraphics[trim={0cm, 0cm, 0cm, 0cm}, clip, width=0.95\linewidth]{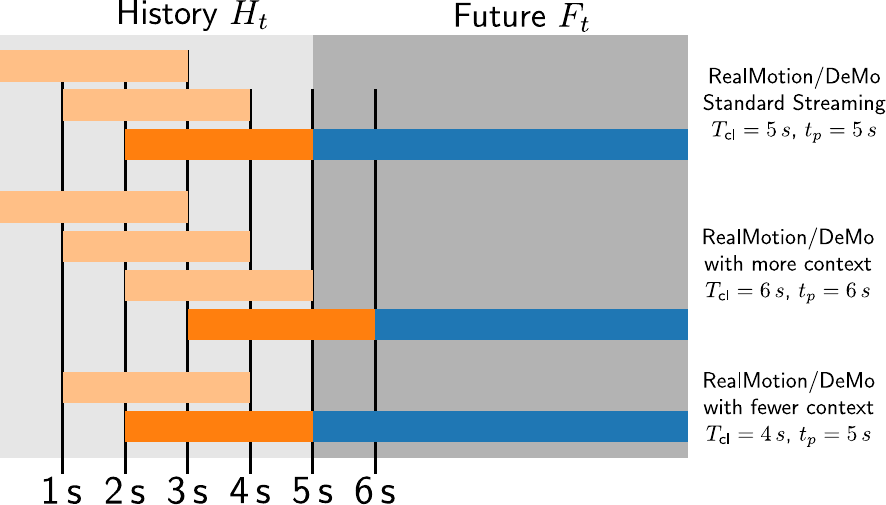} 
    \vspace{-0.2cm}
    \caption{
    Details for evaluating related work on the evolving scene setting (main paper Table 1).
    The first example shows the standard execution on the AV2 benchmark.
    In the second example, we increase the input context by simply executing another streaming step.
    In the last example, we test the models with fewer context by evaluating after two streaming steps.
    }
    \label{fig:supp_timeline_eval}
    \vspace{-0.45cm}
\end{figure}

To evaluate related work on streaming trajectory prediction in the evolving scene setting (main paper Table 1), we benchmark all methods across different prediction time steps \pt~and varying numbers of input windows\cl.
To ensure fair comparison, we keep the observation window (3\,s) and the window displacement (1\,s) identical to the original setup \cite{song2024realmotion, zhang2024demo}, effectively mimicking how these models behave in practical streaming deployment when provided with fewer or more observations.
A visualization of the resulting streaming evaluation protocol is shown in \Cref{fig:supp_timeline_eval}.

\section{Visualizations}
\label{sec:v}
\subsection{Evolving Scenes}
We present qualitative results on scenarios from the Argoverse~2 validation set in \Cref{fig:supp_results}.
Additionally, we provide animated results as separate MP4 files on our project page.
\begin{figure*}[tp]
    \centering
    \resizebox{\linewidth}{!}{
        \begin{tikzpicture}[node distance=0mm, inner sep=0mm]
            \addresrow{0}{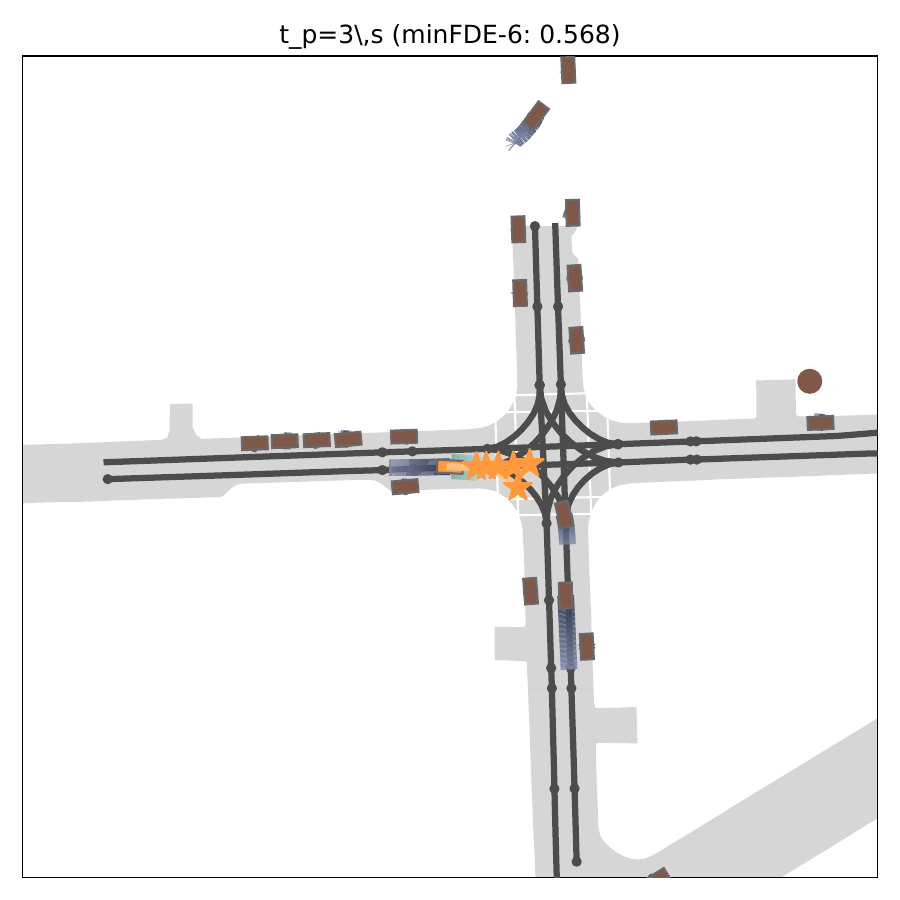}{3}{3}{3}{3}{0.025}{0.01}{}
            \addresrow{1}{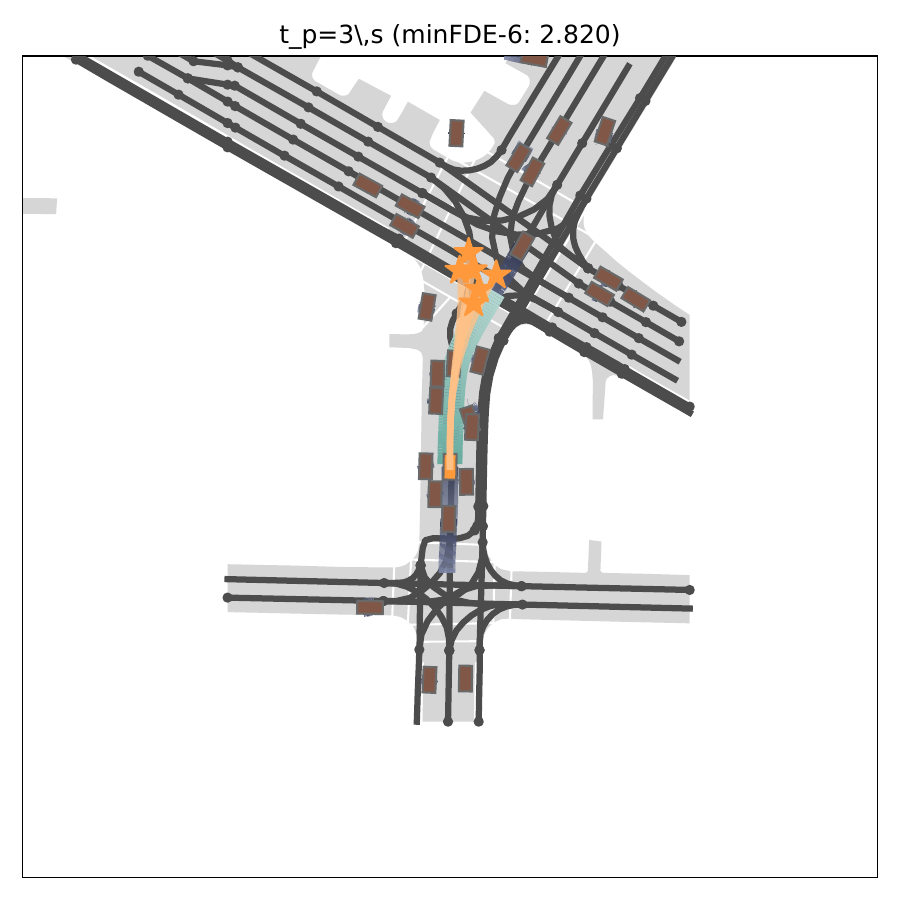}{3}{3}{3}{3}{0.025}{0.01}{0}
            \addresrow{2}{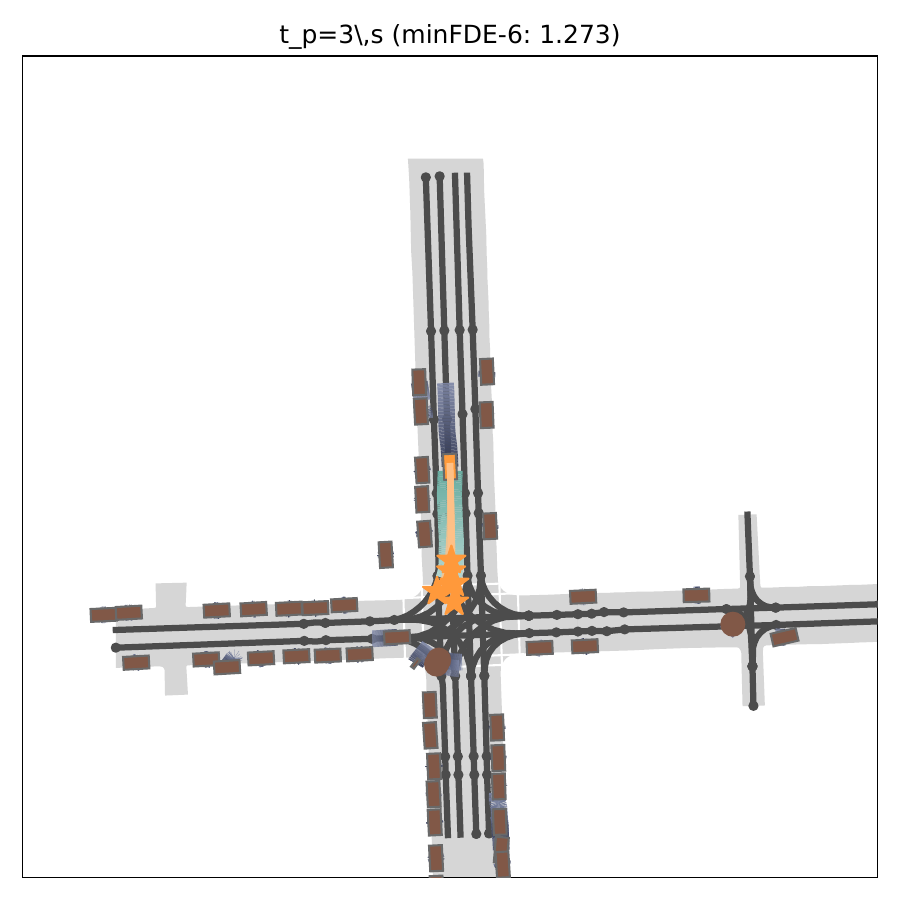}{3}{3}{3}{3}{0.025}{0.01}{1}
        \end{tikzpicture}
    }
    \caption{
    Qualitative single-agent prediction results of our \mn~on scenarios from the Argoverse~2 validation set. We visualize the \textcolor[HTML]{ff9a3a}{\textbf{predictions}} of our streaming-based method at \pt~$\in \{3, 4, 5\}$s.
    The visualizations also show \textcolor[rgb]{0.41, 0.67, 0.63}{\textbf{ground truth future}}, \textcolor[HTML]{384062}{\textbf{historical agent observations}}, and \textcolor[HTML]{815847}{\textbf{surrounding agents}}.
    For comparison, the right column shows the predictions of DeMo~\cite{zhang2024demo} at the AV2 standard prediction time step \pt=5\,s.
}
\label{fig:supp_results}
\end{figure*}

\subsection{Failure Cases}
We present failure cases in which our approach fails to correctly predict future trajectories in \Cref{fig:supp_fail}.
Commonly, failures are introduced by agent movements which cannot be anticipated at the prediction time, often also due to inadequate map data, \eg missing modeling of driveways.
\begin{figure*}[tp]
    \centering
    \resizebox{\linewidth}{!}{
    \begin{tikzpicture}[node distance=0mm, inner sep=0mm]
        \addfailrow{0}{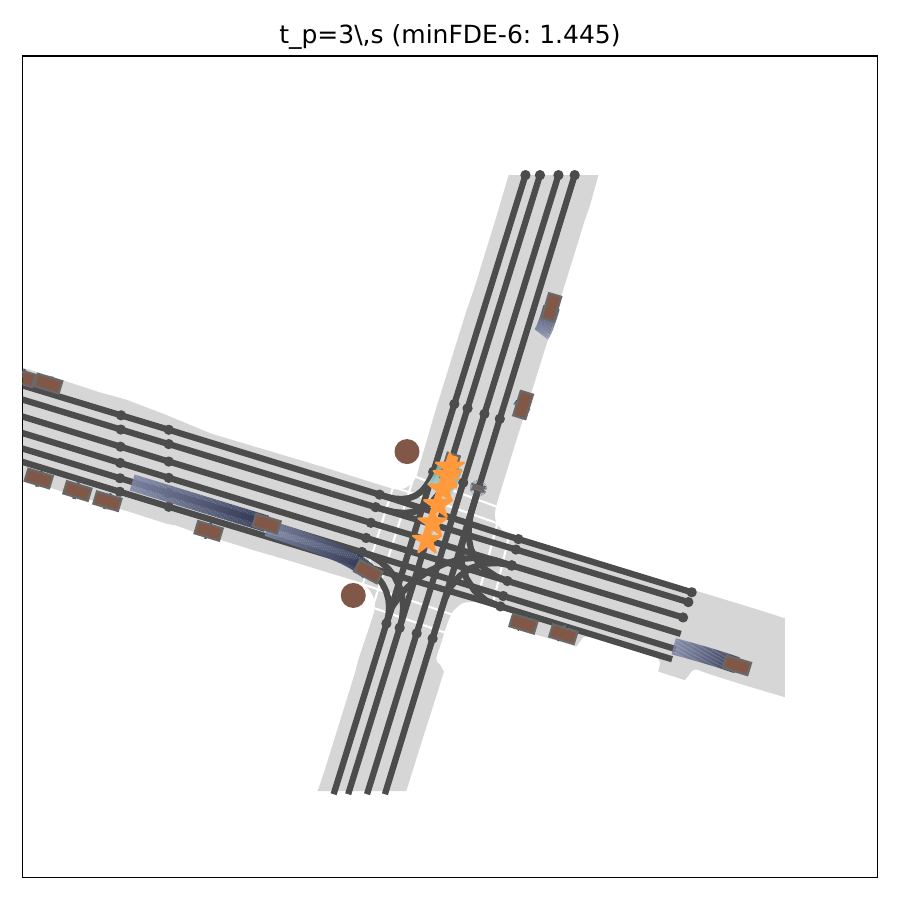}{4}{4}{5}{5}{0.025}{Turn after waiting at intersection}{}
        \addfailrow{1}{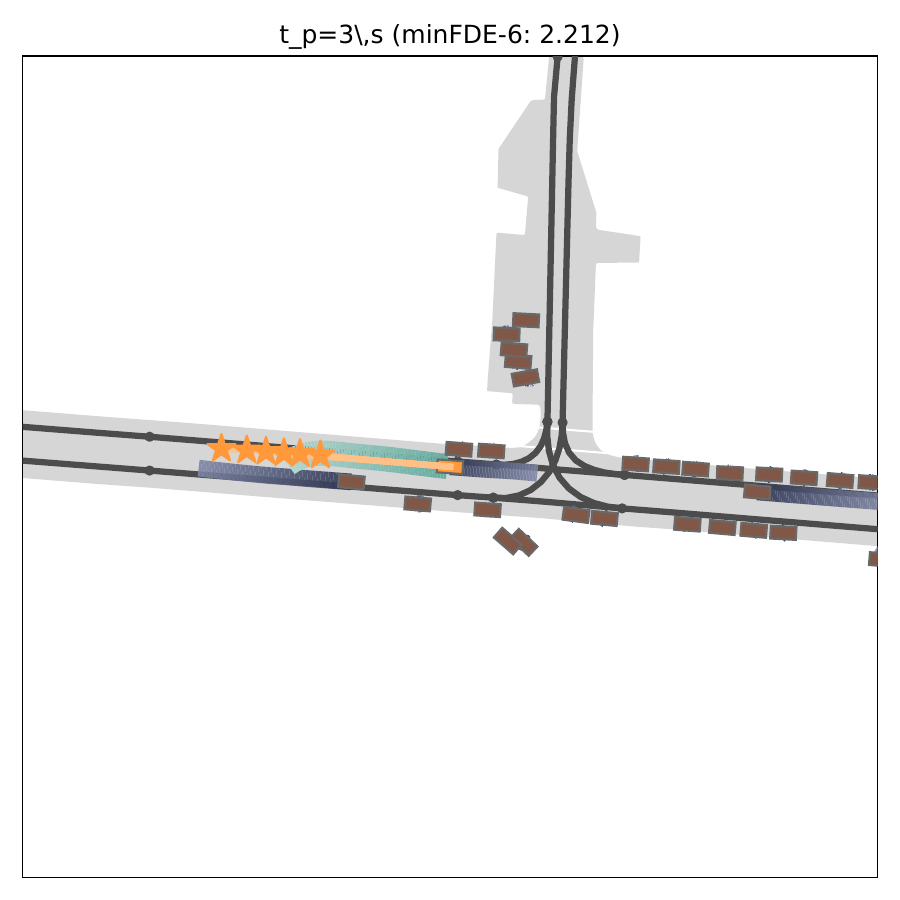}{1}{3}{5}{3}{0.025}{Driveway not mapped}{0}
        \addfailrow{2}{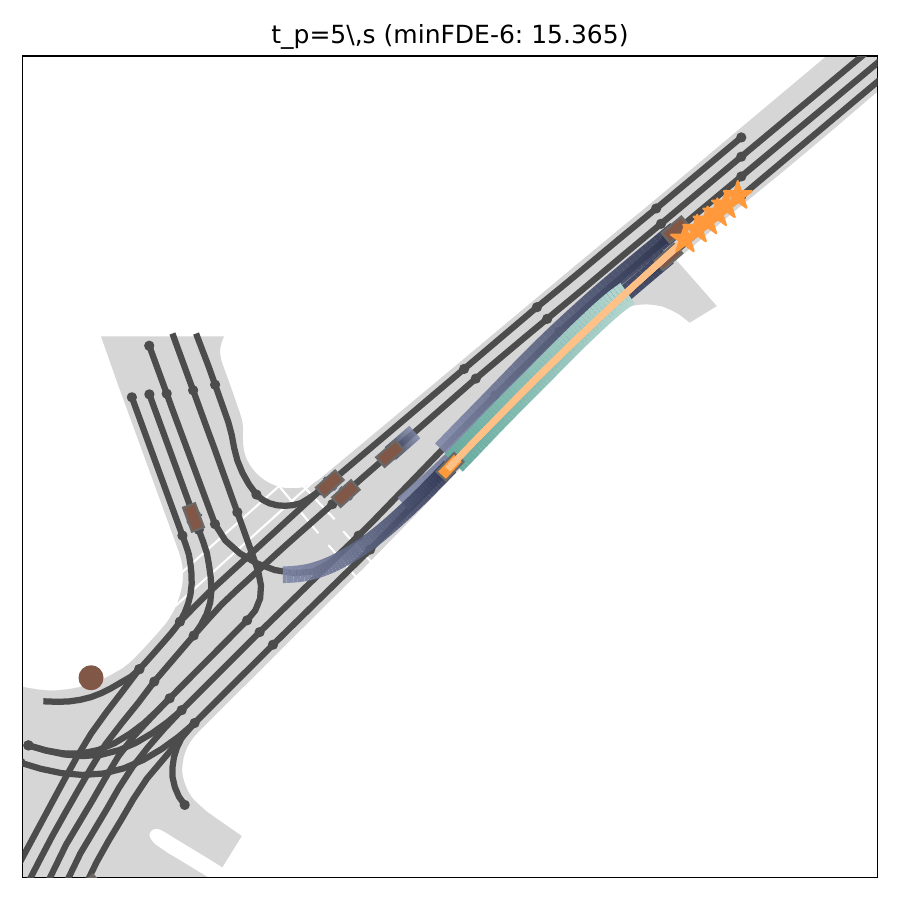}{5}{3}{0.5}{2}{0.025}{Unexpected Stop}{1}        
    \end{tikzpicture}
    }
    \caption{
    Failure cases of our \mn~on scenarios from the Argoverse~2 validation set. 
    We visualize the \textcolor[HTML]{ff9a3a}{\textbf{predictions}} of our streaming-based method at \pt~$\in \{3, 4, 5\}$s.
    The visualizations also show \textcolor[rgb]{0.41, 0.67, 0.63}{\textbf{ground truth future}}, \textcolor[HTML]{384062}{\textbf{historical agent observations}}, and \textcolor[HTML]{815847}{\textbf{surrounding agents}}.
    For comparison, the right column shows the predictions of DeMo~\cite{zhang2024demo} at the AV2 standard prediction time step \pt=5\,s.
    In the first scenario, a vehicle is waiting at an intersection but will initiate a right turn later -- an action that is not anticipated from the current observations.
    In the second scenario, a vehicle turns into a driveway that is not represented in the map data.
    In the final scenario, a vehicle stops in the future for a reason that is not captured in the current perception data.  
    }
    \label{fig:supp_fail}
\end{figure*}

\section{Code Implementation}
\label{sec:ie}
To support reproducibility, we provide our code implementation, including data loaders for all three datasets, as well as the extension to the multi-agent setting.
 \fi

\end{document}